%% file: 0_arxiv_short.tex
\documentclass[11pt]{article}

\usepackage[preprint]{acl}

\usepackage[dvipsnames]{xcolor}
\usepackage{pifont}
\usepackage{times}
\usepackage{latexsym}
\usepackage{graphicx}
\usepackage{adjustbox}
\usepackage{multirow}
\usepackage[normalem]{ulem}
\usepackage{xcolor,colortbl}
\usepackage{soul}
\setuldepth{Berlin}
\usepackage{paralist}
\usepackage{xfrac}
\usepackage[inline]{enumitem}
\usepackage{graphicx}
\usepackage{url}
\usepackage{textcomp}
\usepackage{caption}
\usepackage{subcaption}
\usepackage{float}
\usepackage{placeins}
\usepackage{aliascnt}
\usepackage{mathtools}
\usepackage{amsmath}
\usepackage{hhline}
\usepackage{booktabs}
\usepackage{soul}
\usepackage{bbm}
\usepackage{amssymb}
\usepackage{pifont}
\usepackage{adjustbox}
\usepackage{array}
\usepackage{lipsum}
\usepackage{scrextend}

\usepackage[percent]{overpic}
\usepackage{titlesec}
\usepackage{setspace}
\usepackage{hyperref}
\usepackage{microtype}
\usepackage{footnote}
\usepackage{threeparttable}
\usepackage{tipa}
\usepackage[T1]{fontenc}
\usepackage[utf8]{inputenc}
\usepackage{pgfplots}
\pgfplotsset{compat=1.18}
\usepackage{tikz}
\usetikzlibrary{patterns}

\title{PACUTE: Phonology-, Affix-, and Character-level Understanding\\of Tokens for Filipino}

\author{
 \textbf{Jann Railey Montalan\textsuperscript{1,2}\thanks{Corresponding author: \href{mailto:railey@aisingapore.org}{railey@aisingapore.org}}\thanks{Equal contribution}},
 \textbf{David Demitri Africa\textsuperscript{3}\footnotemark[\value{footnote}]},
 \textbf{Jimson Paulo Layacan},
\\
 \textbf{Richell Isaiah Flores\textsuperscript{4}},
 \textbf{Ivan Yuri De Leon\textsuperscript{4}},
 \textbf{Lance Calvin Gamboa\textsuperscript{5}},
\\
\\
 \textsuperscript{1}AI Singapore,
 \textsuperscript{2}Nanyang Technological University,
 \textsuperscript{3}UK AI Security Institute,
\\
 \textsuperscript{4}Ateneo de Manila University,
 \textsuperscript{5}University of Birmingham
\\
}

\begin{document}
\maketitle
% \footnotetext{
%     \textsuperscript{*} Equal contribution
%     \\Corresponding author: \href{mailto:railey@aisingapore.org}{railey@aisingapore.org}
% }
\begin{abstract}
  Large language models (LLMs) process text as sequences of subword tokens, which can obscure the character-level and morphological structure that underlies word formation. This limitation is most acute for languages with non-concatenative morphology, where standard tokenizers systematically misalign token boundaries with morpheme boundaries. We introduce PACUTE, a diagnostic benchmark of 4,600 tasks designed to evaluate morphological understanding in Filipino, a language characterized by productive infixation, reduplication, and diacritic-driven lexical distinctions that are typically absent from written text. PACUTE includes a hierarchical diagnostic framework of six compositional levels that localizes where morphological understanding breaks down. Evaluating open-weight LLMs and frontier commercial models, we find that open-weight models perform near chance on morpheme decomposition regardless of scale. Frontier models perform much better, often recovering individual affixes under contains-match scoring, but remain far below their character-level ceilings on compositional tasks of morpheme transformations and syllabification. These results identify productive morphological composition, rather than character access alone, as the persistent bottleneck for Filipino word-structure understanding.
\end{abstract}

\section{Introduction}

\begin{figure*}[t]
  \centering
  % Replace with the actual path/filename of your exported React figure
  \includegraphics[width=\textwidth]{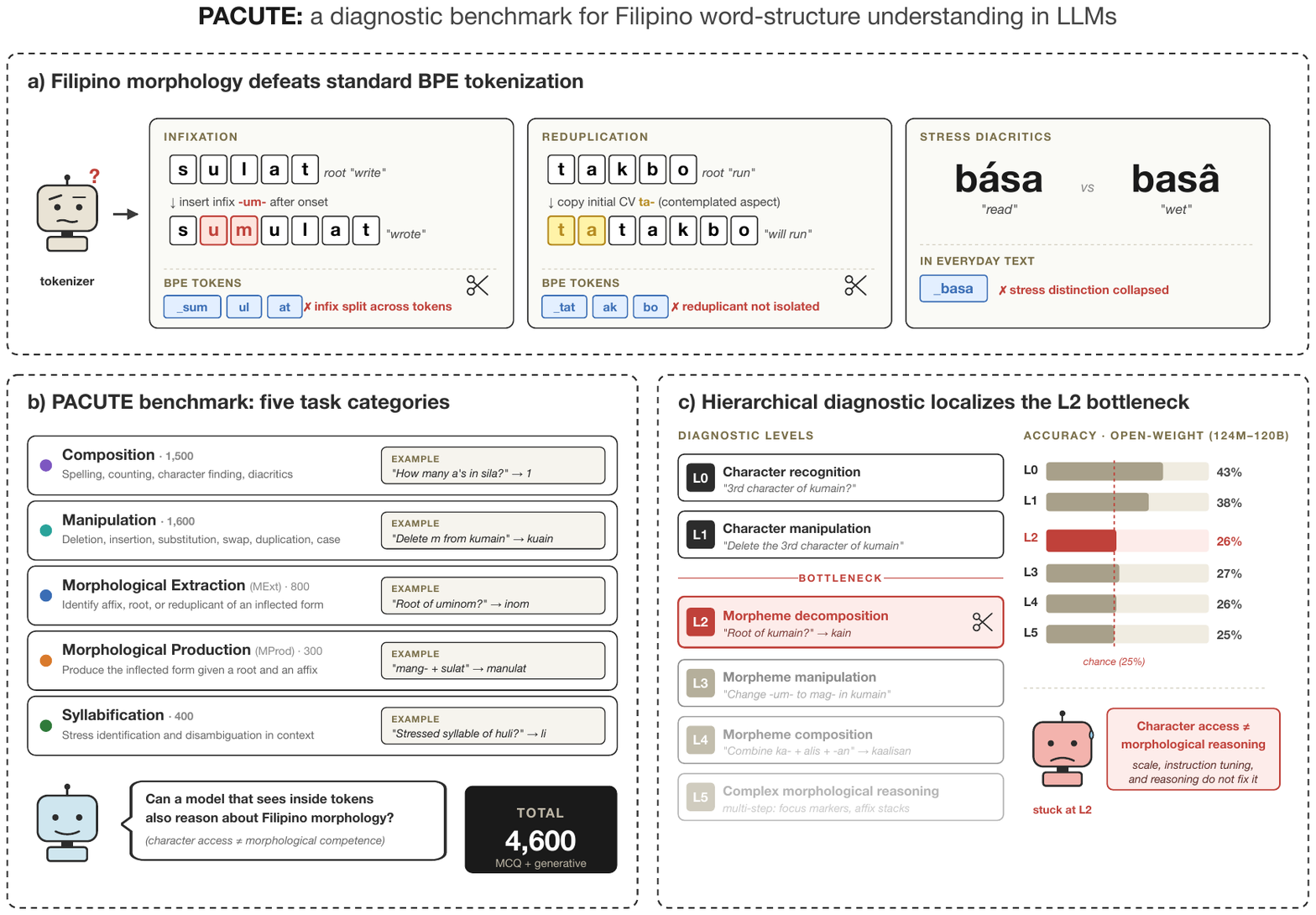}

  \caption{\textbf{Overview of PACUTE.} (A) Filipino morphology poses challenges for standard tokenizers: infixes split roots, reduplication copies syllables, and stress diacritics are typically omitted in text. (B) PACUTE comprises four task categories targeting different levels of word structure understanding. (C) A six-level hierarchical diagnostic localizes failures: models show above-chance performance on character-level tasks (L0–L1) but collapse to chance on morpheme decomposition (L2), with downstream levels inheriting this failure regardless of scale.}
  \label{fig:pacute_overview}
\end{figure*}

Large language models segment text into subword tokens and process them as atomic units, which limits their direct access to character-level structure. Benchmarks such as CUTE \citep{edman-etal-2024-cute} and LangGame \citep{sims2025stochastokimprovingfinegrainedsubword} demonstrate that many LLMs struggle with basic token-level tasks such as counting characters, character manipulations, and substring detection.

\paragraph{Filipino poses distinct challenges for tokenizers.} We claim that Filipino, the national language of the Philippines, is a natural (and naturalistic) testbed to evaluate such capabilities in LLMs, for two reasons.

\paragraph{Productive non-concatenative morphology.} Filipino uses infixes (e.g., -\textit{um}-, -\textit{in}-) that split root words and partial reduplication of initial syllables. For example, the root \textit{kain} ``eat'' becomes \textit{kumain} ``ate'' via the actor-focus infix -\textit{um}-, inserted after the first consonant. Standard byte-pair encoding tokenizers \cite{sennrich-etal-2016-neural} are suboptimal for language model pre-training \cite{bostrom2020byte}, and frequently segment such words at arbitrary boundaries that ignore morpheme structure.

\paragraph{Omitted stress and glottal markers.} Filipino orthography can encode lexical distinctions through diacritics: acute (pahilís), grave (paiwà), and circumflex (pakupyâ). These create minimal pairs \citep{warstadt2020blimp}: \textit{bása} ``read'' vs. \textit{basâ} ``wet''; \textit{súka} ``vomit'' vs. \textit{sukà} ``vinegar''; \textit{táyo} ``we'' vs. \textit{tayô} ``stand''. However, everyday Filipino text typically omits these markers, creating systematic lexical ambiguity invisible to models.

Based on this, we introduce PACUTE (\textbf{P}honology-, \textbf{A}ffix-, and \textbf{C}haracter-level \textbf{U}nderstanding of \textbf{T}okens \textbf{E}valuation), a benchmark of 4,600 synthetic tasks targeting Filipino composition, manipulation, morphological extraction, morphological production, and syllabification, each available in multiple-choice (log-probability) and generative formats. PACUTE also includes a hierarchical diagnostic set (six levels) that separates prerequisite character skills from morpheme-level operations, enabling localization of bottlenecks.

We evaluate several pre-trained and instruction-tuned LLMs across major open families and sizes, as well as run continued pre-training on Gemma-2-2B with three preprocessing/tokenization regimes: vanilla BPE, stochastic token expansion (StochasTok) \cite{sims2025stochastokimprovingfinegrainedsubword}, and a Filipino morphology-aware expand/contract method based on StochasTok. Our results show: \textbf{(i) open-weight models collapse to chance on morpheme decomposition} under MCQ log-probability scoring, despite above-chance performance on lower-level character tasks; and \textbf{(ii) frontier commercial models} often recover individual affixes, but \textbf{remain substantially below their character-level ceilings on hierarchical morpheme transformations and syllabification}; Filipino-relevant pre-training improves mid-range models, while tokenization interventions alone produce limited gains and can induce catastrophic forgetting.

\section{Related Work}
% CUTE evaluates LLMs’ token composition knowledge via spelling, orthographic similarity, and character-level manipulation. LangGame-type tasks probe letter counts, prefix/suffix/substring identification, and length comparisons. Stochastic tokenization (e.g., BPE-dropout, and recent methods that probabilistically split tokens during training) has been proposed to expose subtoken structure and improve performance on such tasks.

% Morphology benchmarks (e.g., SIGMORPHON inflection tasks) target generalization in morphosyntax, but typically rely on overt segmental changes in languages with transparent orthography. Filipino poses unique orthography–morphology interactions such as infixation and reduplication crosscut subword boundaries, and diacritics that encode stress and final glottal stops that are often omitted in natural text.

% Our work complements these lines by designing Filipino-specific subword tasks that explicitly require handling infixes, reduplication, morphophonemic alternations, and stress.

\paragraph{Relevant properties and evaluation of Filipino.} Filipino has been characterized in the linguistic literature as exhibiting several non-concatenative morphological processes, including infixation (e.g., \textit{-um-}, \textit{-in-}), partial and full reduplication, and distinctions in stress and glottal stop that may be encoded through diacritics \cite{schachter1983tagalog, yap1967synchronic}. These phenomena interact closely with phonological structure and are not always transparently reflected in surface orthography, particularly in naturally occurring text where diacritics are frequently omitted. As a result, many linguistically meaningful contrasts in Filipino are obscured at the character and token levels typically used in computational models.

Despite these properties, computational research on Filipino has largely focused on downstream NLP tasks such as sentiment analysis, part-of-speech tagging, and named entity recognition \cite{cruz2022improving, visperas2023itanong}. A recent benchmark, BATAYAN \cite{montalan2025batayan}, broadens evaluation coverage across understanding and generation tasks, yet remains primarily task-oriented rather than diagnostic of Filipino-specific linguistic structure.

\paragraph{Token and character level evaluations.} In the broader literature, several studies have introduced token- and character-level diagnostic evaluations, for other languages. These include tests based on spelling, orthographic similarity, and character-level perturbations \cite{itzhak2022models, huang2023inducing}, as well as synthetic tasks involving letter counting, prefix or suffix identification, and substring detection \cite{kaushal2022tokens, efrat2023lmentry}. CUTE extends this line of work by systematically evaluating models’ token composition knowledge through character-level manipulations and token-based reasoning tasks \cite{edman-etal-2024-cute}. While effective as a general diagnostic framework, CUTE is designed to be language-agnostic and does not explicitly target non-concatenative morphology or phonology-driven alternations.

Morphology benchmarks such as SIGMORPHON evaluate inflection across diverse languages \cite{vylomova2020sigmorphon}, but emphasize concatenative processes and transparent orthography, which are assumptions that do not hold for Filipino. Subword tokenization variants such as BPE-dropout \cite{provilkov-etal-2020-bpe} improve robustness to surface variation, but evaluation has focused on downstream performance rather than linguistically grounded morphological structure. PACUTE fills this gap with Filipino-specific diagnostics targeting phonology, affixation, and character-level structure.

% incorporate something about stochastok patok

\section{PACUTE: Task Suite}
\label{sec:pacute}

\paragraph{Goal.}
PACUTE is a diagnostic benchmark for Filipino word-structure competence: whether a model can (i) access character-level information inside tokens and (ii) use that information to reason about productive morphology (especially infixation and reduplication). PACUTE is not designed mainly to measure general Filipino ``understanding’’ via downstream tasks; rather, it targets concrete, local operations that underlie morphological generalization.

\paragraph{Why tokenization is a plausible failure source.}
Standard subword tokenizers (e.g., BPE) optimize compression and frequency statistics rather than morpheme boundary preservation. For Filipino, infixes and reduplicated segments often fall inside token interiors or are split inconsistently across tokens, making them hard to access as reusable units during pre-training and inference. Diacritics encoding lexical stress distinctions (\textit{bása} ``read’’ vs.\ \textit{basâ} ``wet’’) are nearly always absent from digital text \cite{almario2014masinop}, so models face a systematic mismatch between lexicographic and naturally occurring forms. Together, infixation, reduplication, morphophonemic alternations, and diacritic omission create a setting where surface-form competence is not sufficient for morphological competence (see Appendix~\ref{sec:filipino-morphology} for a detailed treatment). This motivates two PACUTE design choices: (i) tasks that explicitly require operations aligned with morphological structure rather than only generic character probes, and (ii) a hierarchical diagnostic that separates prerequisite character skills from morpheme-level skills, allowing localization of where the compositional pipeline breaks down.

\paragraph{Task formats.}
Each PACUTE task is provided in two evaluation formats. \textbf{MCQ} instances present four answer options (chance = 25\%). \textbf{GEN} instances require the model to produce an answer string. We include both formats because MCQ reduces generation and formatting confounds, while GEN tests whether the model can reliably execute the operation. Scoring procedures for both formats are described in \S\ref{sec:evaluation}.

\subsection{Main suite and coverage}
The main PACUTE suite contains 4,600 tasks across five categories, each targeting a different slice of subtoken competence:
(i) \textbf{Composition} (950 MCQ + 550 GEN),
(ii) \textbf{Manipulation} (800 MCQ + 800 GEN),
(iii) \textbf{Morphological extraction} (400 MCQ + 400 GEN),
(iv) \textbf{Morphological production} (150 MCQ + 150 GEN),
(v) \textbf{Syllabification} (200 MCQ + 200 GEN).
The composition and manipulation tasks provide character-level controls; the morphological extraction, morphological production, and syllabification tasks target Filipino-specific structure.

\paragraph{Composition tasks.}
Composition tasks test whether a model can use orthographic information that is typically hidden by tokenization: character counting, identification of specific characters/diacritics, and simple string properties. These are Filipino-adapted variants of standard character-level probes, using Filipino lexicon and (where relevant) diacritic-bearing forms. The intent is to separate ``can the model see inside tokens?'' from morphology-specific reasoning.

\paragraph{Manipulation tasks.}
Manipulation tasks test deterministic character operations such as insertion, deletion, substitution, and swapping. These tasks are intentionally low-level: they test whether models can execute controlled edits that mirror the mechanics needed for infix insertion and reduplication (even if the correct position for the edit is linguistically determined in the affixation tasks). This category also functions as a sanity check: if a model cannot reliably delete or insert characters, failures on morphology may be uninterpretable.

\paragraph{Morphological extraction and production tasks.}
Morphological tasks test whether models can identify and apply Filipino affixes, including prefixes, suffixes, and crucially, infixes. \textit{Extraction} asks which affix, reduplicant, or root can be identified from an inflected form (e.g., word \textit{kumain} $\rightarrow$ root \textit{kain} $+$ completed aspect infix \textit{-um-}). \textit{Production} asks for the correct derived form given a root and affix (e.g., apply completed aspect affix \textit{-in-} to \textit{sulat} $\rightarrow$ \textit{sinulat}). 
% Distractors are constructed to be plausible completions (e.g., competing Filipino affixes) for the task (as opposed to, say, arbitrary strings).

\paragraph{Syllabification tasks.}
Syllabification tasks test the phonological scaffolding needed for Filipino morphology. We treat syllable-level competence as an intermediate level between raw character manipulation and morpheme-level reasoning. This category includes stress identification tasks, which ask the model to identify the stressed syllable of a given word as it appears in a disambiguating sentence context. Because Filipino orthography conventionally omits stress diacritics, stress assignment is non-trivial: many words are orthographically identical but phonologically and semantically distinct depending on which syllable carries primary stress (e.g., \textit{sála} ``sin'' vs. \textit{salà} ``filter'' vs. \textit{salâ} ``broken''). Models must therefore use both lexical knowledge and sentential context to resolve stress.

\paragraph{Hierarchical diagnostic benchmark.}
In addition to the main suite, PACUTE includes a hierarchical diagnostic benchmark (600 MCQ + 600 GEN) organized into six levels:
Level 0 (character recognition),
Level 1 (character manipulation),
Level 2 (morpheme decomposition),
Level 3 (morpheme manipulation),
Level 4 (morpheme composition),
Level 5 (complex multi-step transformations).
The hierarchy is designed to localize failures: Levels 3--5 presuppose that the model can decompose a word into morphemes (Level 2). Interpretation of per-level results is discussed in \S\ref{sec:evaluation}.

\paragraph{Prompting and instance structure.}
All instances use short, uniform templates with minimal instruction overhead. GEN instances use a structured XML format with a \texttt{<reflection>} reasoning trace followed by an \texttt{<answer>} block; the evaluator extracts only the answer before scoring. Full prompting details, chat-template injection, and thinking-mode handling are described in Appendix~\ref{sec:prompting}.

\section{Data Construction}
% We synthesize task instances from public resources and deterministic rules to avoid contamination from proprietary pre-training corpora and large language model slop (cite anisotropy paper and LLM vocab collapse paper)

\label{sec:data}
All PACUTE tasks are generated deterministically from public lexical resources and hand-curated annotations, without any use of generative models for content production.\footnote{Code and benchmark data can be found at \url{https://github.com/raileymontalan/pacute-bench}.}
% \footnote{Code and benchmark data will be made publicly available upon acceptance.}
This design choice is motivated by two considerations: (i) avoiding contamination of the benchmark with patterns already present in LLM pre-training corpora, and (ii) ensuring that every instance has a verifiable gold answer derived from linguistic rules rather than model-generated text \cite{bender-etal-2021-parrots}.

\begin{table}[t]
  \centering
  \setlength{\tabcolsep}{4pt}
  \caption{PACUTE task statistics by category and format. \textit{LangGame} = language-agnostic control; \textit{MDA (multi-digit addition)} = non-linguistic control; \textit{CUTE} = character-understanding control (generative only).}
  \label{tab:task-stats}
  % \scriptsize
  \small
  \begin{tabular}{lrrr}
    \toprule
    \textbf{Category}        & \textbf{MCQ} & \textbf{GEN} & \textbf{Total}    \\
    \midrule
    \multicolumn{4}{l}{\textit{Main suite}}                                    \\
    \quad Composition        & 950          & 550          & 1{,}500           \\
    \quad Manipulation       & 800          & 800          & 1{,}600           \\
    \quad Morph.\ extraction & 400          & 400          & 800               \\
    \quad Morph.\ production & 150          & 150          & 300               \\
    \quad Syllabification    & 200          & 200          & 400               \\
    \multicolumn{4}{l}{\textit{Hierarchical (L0--L5, 100 MCQ+GEN each)}}       \\
    \quad Total hierarchical & 600          & 600          & 1{,}200           \\
    \multicolumn{4}{l}{\textit{Controls}}                                      \\
    \quad LangGame / MDA     & 1{,}000      & 1{,}000      & 2{,}000           \\
    \quad CUTE               & ---          & 1{,}400      & 1{,}400           \\
    \midrule
    \textbf{Total}           &              &              & \textbf{9{,}200} \\
    \bottomrule
  \end{tabular}
\end{table}

\paragraph{Lexical resources.}
The primary source is a syllabified word list from the \textit{UP Diksiyonaryong Filipino} \cite{updiksiyonaryo2001} (16,828 entries with syllable boundaries, stress, and POS), paired with a Filipino corpus frequency list (118,801 forms) for frequency-weighted sampling. Morphological tasks draw from a manually curated dataset of inflection (e.g., \textit{takbo} $+$ \textit{-um-} $\rightarrow$ \textit{tumakbo}), assimilation (e.g., \textit{naNG-} $+$ \textit{bigay} $\rightarrow$ \textit{namigay}), and reduplication patterns \cite{zamar2022filipino, schachter1983tagalog}. Syllabification tasks use a handcrafted corpus of minimally contrastive word pairs in disambiguating sentence context (e.g., \textit{búhay} ``life'' vs.\ \textit{buháy} ``alive''). All task instances are generated by a deterministic Python pipeline; items are available in English and Filipino prompt formats and summarized in Table~\ref{tab:task-stats}.

\section{Evaluation}
\label{sec:evaluation}

\paragraph{Overview.}
We use PACUTE to evaluate (i) a broad set of LLMs (zero-shot) and (ii) a controlled continued-pre-training (CPT) study where we vary tokenization during Filipino-domain adaptation. We report normalized accuracy and F1 score on MCQ tasks (4-way; chance = 25\%) and string-match accuracy on generative tasks.

\paragraph{Models.}
We evaluate open-weight models spanning several major families (Gemma, GPT-2, GPT-OSS, Llama, Mistral, Phi, Qwen, SEA-LION, DeepSeek, Kimi), parameter scales from 124M to 1T, and both base pre-trained, instruction-tuned, and thinking variants. We additionally evaluate frontier commercial models (Claude 4 series, GPT-5 series, and Gemini 3 series) on generative benchmarks only (commercial APIs for most models do not expose token log-probabilities for MCQ scoring). This mix allows us to separate effects of scale, instruction tuning, regional pre-training data (SEA-LION), and frontier capability.

\paragraph{Scoring.}
All models are evaluated on the full PACUTE suite, hierarchical benchmark, and controls (CUTE, LangGame, MDA). For MCQ, we select the option with highest conditional log-probability and report normalized accuracy and F1 (4-way; chance = 25\%). For GEN, we use exact match after lightweight normalization (case, whitespace, leading/trailing hyphens on affix strings) and contains-match for tasks where models may produce extra context. Per-level hierarchical results (L0--L5) are interpreted diagnostically: if L2 (morpheme decomposition) is at chance, L3--L5 are expected to collapse regardless of L0--L1 performance.

\paragraph{Continued pre-training (CPT) setup.}
For CPT, we adapt Gemma-2-2B on the SEA-PILE v2 Filipino corpus (7.4GB) for up to 5,000 steps, saving checkpoints every 1,000 steps. We compare three segmentation regimes applied at preprocessing time while keeping the underlying model architecture fixed:
(i) vanilla (unchanged BPE tokenization),
(ii) StochasTok (stochastic token expansion; $\sim$10\% split rate),
(iii) Patok (morphology-aware expand/contract using Filipino affix rules).
We evaluate each checkpoint on the same benchmark suite.

\paragraph{Human baseline.}
We establish a human baseline on a PACUTE subset using three native Filipino speakers, reporting average accuracy and Fleiss'~$\kappa$ for inter-annotator agreement (Appendix~\ref{sec:evaluation_results}).

\section{Results}
\label{sec:results}

\input{fig_results}

Full numerical results are reported in Tables~\ref{tab:human-baselines-mcq},~\ref{tab:human-baselines-gen},~\ref{tab:results-pt},~\ref{tab:results-it:mcq},~\ref{tab:results-it:gen},~\ref{tab:results-commercial},~\ref{tab:cpt-results}, and in the Appendix. Figure~\ref{fig:results_by_category} summarizes the main pattern: strong models approach ceiling on character-level and language-agnostic controls, but remain substantially weaker on Filipino-specific phonological and morphological tasks. The gap is clearest on the Hierarchical benchmark and on Syllabification, which require models to use morpheme structure, syllable structure, stress, and phonologically conditioned transformations rather than simple character edits.

\paragraph{Morpheme decomposition is difficult for open-weight models.}
Among open-weight models evaluated with MCQ log-probability scoring, performance on the Hierarchical benchmark remains close to chance. Normalized accuracy ranges from $-3.6$ to $+2.9$ across 32 open-weight variants (mean: $-0.6$). This collapse is driven by Level~2 morpheme decomposition: models often fail to identify the root and affix structure of Filipino words, even when they perform above chance on lower-level character recognition and manipulation tasks. Scale alone does not remove this failure. Larger open-weight models, including Gemma-4-31B and Qwen-3.6-27B, do not reliably outperform much smaller models on morpheme decomposition.

\paragraph{Character-level controls are largely solved by frontier models.}
Commercial models perform near ceiling on the easiest controls: MDA (99.4--100.0\% CM), LangGame (94.9--100.0\%), and Manipulation (GPT-5.5: 100.0\%, DeepSeek-R1: 96.9\% CM). These results confirm that PACUTE's harder tasks reflect genuine morphological difficulty, not generic instruction-following failure.

\paragraph{Morphological extraction is recoverable for strong models, but remains formatting-sensitive.}
Performance on Morphological Extraction is consistently higher under contains-match than exact match (e.g., Gemini-3.5-Flash: 89.5\% EM / 90.5\% CM; Claude-Sonnet-4.6: 74.0\% / 76.2\%), indicating models often recover the correct affix but not always in the required form.

\paragraph{Hierarchical transformations remain the central bottleneck.}
The Hierarchical benchmark remains substantially below the character-level ceiling. The best model, GPT-5.5, reaches 64.7\% EM and 77.5\% CM on Hierarchical-GEN, compared with 100.0\% CM on Manipulation-GEN. DeepSeek-R1 reaches 72.7\% CM, Kimi-K2-Thinking reaches 67.2\%, Claude Sonnet~4.6 reaches 68.0\%, and GPT-5-Mini reaches 67.7\%. This pattern suggests that the difficulty is not identifying individual characters or even recovering individual affixes, but composing several operations: decomposing a word into morphemes, manipulating those morphemes, and recombining them into the correct surface form. PACUTE therefore isolates a gap between local subtoken access and productive morphological reasoning.

\paragraph{Syllabification exposes a separate phonological weakness.}
Syllabification is difficult even for frontier models. The best performers are GPT-5.5 (83.0\% CM) and Gemini-3.5-Flash (82.5\%), while several strong models score much lower (GPT-5-Mini: 46.0\%; Gemini-3.1-Flash-Lite: 38.0\%). Even models that handle character manipulation well may still lack stable representations of Filipino syllable structure, stress, and the \textit{ng} digraph.

\paragraph{Reasoning models narrow but do not close the gap.}
Thinking models are often strong, but their gains are uneven. DeepSeek-R1 obtains one of the highest Hierarchical scores among commercial models (72.7\% CM), and Kimi-K2-Thinking also performs competitively (67.2\% CM). Gemini-3.5-Flash is especially strong on Morphological Extraction, Morphological Production, and Syllabification, reaching 90.5\%, 90.0\%, and 82.5\% CM respectively, but remains lower on Hierarchical at 62.2\% CM. These results suggest that reasoning can help when the relevant linguistic knowledge is available, but it does not by itself guarantee robust morpheme decomposition or composition.

\paragraph{Pre-training data matters below the frontier.}
Among open-weight models, SEA-LION models trained on Southeast Asian data outperform generic models of comparable scale on Filipino-specific tasks. SEA-LION-Qwen-v4.5-27B achieves 85.8 normalized accuracy on MProd-MCQ and 27.5 on Manip-MCQ, compared with 83.1 and 26.2 for Qwen-3.6-27B. This suggests that Filipino-relevant pre-training data improves mid-range model performance, even though frontier commercial models can partially compensate through general capability and stronger instruction following.

\paragraph{Human baseline.}
Human performance is substantially above model performance on all categories. Humans particularly outperform the strongest models on MExt-MCQ (0.925 vs. 0.895 EM, MProd-GEN (0.911 vs. 0.900 EM), and Syll-GEN (0.967 vs. 0.830 EM). Overall agreement among annotators for PACUTE is high (90.4\% for GEN, 85.0\% for MCQ). Moreover, Fleiss' $\kappa$ of 85.5\% for MCQ and 74.0\% for GEN indicates almost perfect and substantial agreement respectively.

\paragraph{Summary.}
PACUTE reveals a hierarchy of competence: strong models solve character edits and language-agnostic controls, and often recover individual affixes, but remain well below their character-level ceilings on Hierarchical and Syllabification tasks. Scale, instruction tuning, and reasoning all help, but productive morphological composition (such as decomposing words, tracking syllable structure, applying phonologically conditioned transformations) remains the persistent bottleneck.

\section{Error Analysis}
\label{sec:error-analysis}
We conduct a manual error analysis on sampled GEN responses from instruction-tuned models, focusing on GEN because reasoning traces and surface-form choices are visible there. We identify four recurring error types.

\paragraph{Type I: Instruction following failure.}
Many GEN errors reflect non-compliance rather than a linguistic gap: generating a full sentence instead of a word, explaining instead of answering, or responding in English to a Filipino prompt. Thinking-mode models are especially prone, as the correct answer often appears mid-chain but is not surfaced as the final response. These errors inflate raw GEN error rates and motivate contains-match scoring.

\paragraph{Type II: Linguistic error.}
Models attempt the correct operation but apply rules incorrectly. The most frequent subtype is infix placement error: the infix is identified (e.g., \textit{-um-}) but inserted at the wrong position (e.g., \textit{*sulumat} instead of \textit{sumulat}). Other subtypes include root boundary confusion, \textit{ng}-digraph mishandling in syllabification, and reduplication copying at the wrong edge.

\paragraph{Type III: Reasoning inconsistency.}
Thinking models (e.g., Qwen-3 variants) sometimes reach the correct intermediate conclusion in the scratchpad yet output a contradicting final answer, such as by correctly counting syllables then rationalizing a different number, or identifying the right root/affix then selecting a wrong MCQ option. This suggests the final answer is not robustly conditioned on the chain-of-thought.

\paragraph{Type IV: Systematic model bias.}
The most distinctive error class is a \textbf{penultimate-stress bias}: models default to labeling the second-to-last syllable as stressed regardless of actual diacritic position. Analogous biases appear in affix selection (preference for high-frequency \textit{mag-} even in illicit contexts). Neither scale nor extended reasoning corrects these biases. Models also fail to use sentential context for stress disambiguation---e.g., Gemma-3-27B-IT answers \textit{ga} for \textit{galing} in all contexts even when the sentence selects the noun reading requiring final stress---and treat the digraph \textit{ng} as two Unicode characters, causing systematic syllable overcounting.

\section{Discussion}
\label{sec:discussion}

\paragraph{Morpheme decomposition as a training data problem.}
The near-random performance on Level~2 decomposition tasks among open-weight models points to a missing-knowledge failure rather than an insufficient-capacity failure. Frontier commercial models partially overcome this floor, but their residual errors concentrate specifically on phonologically conditioned processes---nasal assimilation, vowel-initial root alternations---that require productive rule application rather than memorization. This suggests two stacked bottlenecks: (i) Filipino morphological structure is underrepresented in pre-training corpora of English-dominant models, and (ii) even when models acquire partial affix knowledge at scale, they fail to learn the phonological rules governing affix--root interactions. BPE tokenization compounds both by fragmenting infixed and reduplicated forms at boundaries that obscure morpheme structure. Composition tasks probing diacritic identification score notably lower than non-diacritic character counting for most models, consistent with stress and glottal markers being nearly absent from everyday Filipino text \cite{almario2014masinop}---a gap with downstream consequences for stress disambiguation and lexical sense selection invisible to task-oriented benchmarks.

\paragraph{Why tokenization-based interventions have limited impact.}
  The CPT results with StochasTok and Patok (Appendix \ref{sec:cpt-results}) show that all three regimes, including vanilla BPE,
  underperform the un-CPT'd base on the PACUTE main-suite. Crucially,
   Hierarchical accuracy remains pinned at the chance floor
  for all three regimes (21.2--24.7\%), confirming that L2
  morpheme decomposition is unmoved by either stochastic
  token expansion or morphology-aware segmentation. This is
  consistent with prior work showing that BPE-dropout and
  similar methods improve robustness to surface variation
  but do not inject morphological abstraction when the
  underlying training signal is absent
  \cite{bostrom2020byte,
  sims2025stochastokimprovingfinegrainedsubword}. None of these CPT setups
  yield a net gain for Filipino morphology, and simply
  exposing the model to more Filipino text under alternative
   segmentation schemes is insufficient without mechanisms
  that preserve general capabilities or explicitly align
  token boundaries with morpheme boundaries at scale.
  
\paragraph{Implications for low-resource morphologically complex languages.}
PACUTE's diagnostic hierarchy provides a template for identifying where model competence breaks down for languages with non-concatenative morphology. The L0/L1 scores show that models can access character-level information inside tokens for Filipino to a limited degree; the abrupt collapse at L2 precisely locates the gap. This has practical implications: downstream Filipino NLP tasks that rely on aspect or voice distinctions (which are morphologically encoded) should not be expected to benefit from scale alone if morpheme decomposition is absent. The SEA-LION advantage suggests that domain-specific pre-training on high-quality Filipino text is the most actionable intervention currently available.

\section*{Limitations}
PACUTE evaluates word-level morphological and character-level competence in a controlled, synthetic setting. Several limitations should be noted. 

\paragraph{Lexical resource coverage.} All tasks are generated from a curated lexical resource and a manually annotated affix set. The word pool (16,828 entries from the \textit{UP Diksiyonaryong Filipino}) does not cover informal registers, code-switching, or borrowings from Spanish and English, which are pervasive in natural Filipino text; infix instances in particular are underrepresented in the affixation task set, limiting the diagnostic power of infix-specific categories in the hierarchical benchmark. 

\paragraph{Transfer to naturalistic tasks.} Model performance on PACUTE may therefore not transfer directly to naturalistic Filipino NLP tasks. Tasks that involve diacritic-bearing forms assume that models have been exposed to or can infer stress distinctions. However, diacritics are nearly always absent from digital Filipino text, meaning that performance on diacritic tasks may underestimate competence that a model could demonstrate if given diacritic-marked input.

\paragraph{Reproducibility of continued pretraining.} The Gemma-based experiments were conducted on an internal compute cluster using the NVIDIA NeMo training framework for continued pretraining; we do not release the continued pretraining code to avoid disclosing proprietary infrastructure details, though all pretraining runs used NeMo without custom modifications to the training loop or data pipeline.

% \textit{Evaluation only.} PACUTE provides no training split. It is designed for zero-shot or few-shot evaluation; the benchmark should not be used to train models directly, as this would conflate evaluation with optimization.

\section*{Ethical Considerations}

\paragraph{Data sources and licensing.} All lexical data are drawn from published reference materials, namely the syllabified word list from the \textit{UP Diksiyonaryong Filipino} \cite{updiksiyonaryo2001}, morphological patterns from \cite{zamar2022filipino} and \cite{schachter1983tagalog}, and from the SEA-PILE-v2 Filipino text corpus \cite{ng2025sealionsoutheastasianlanguages}. No personally identifiable information, social media data, or sensitive personal communications were used in task construction. SEA-PILE v2 is released under the ODC-By 1.0 license; we use it in accordance with its terms. The \textit{UP Diksiyonaryong Filipino} is a copyrighted reference work published by the University of the Philippines. 

\paragraph{Intended use.} PACUTE is intended as a diagnostic tool for identifying specific morphological competence gaps in LLMs, not as a general-purpose measure of Filipino language ability. PACUTE should not be used to rank models for production deployment in Filipino-language applications without considering downstream task performance. 

\paragraph{Sociolinguistic scope.} Filipino is the national language of the Philippines. We acknowledge that the computational treatment of Filipino morphology in this work is primarily from a formal linguistic perspective and may not capture the full range of dialectal variation or the sociolinguistic complexity of language use in the Philippines. We encourage follow-up work that extends PACUTE-style diagnostics to other Philippine languages (e.g., Cebuano, Ilocano) and to more naturalistic evaluation settings.

\paragraph{Use of AI assistants.} AI language models (Claude) were used for proofreading and editing of manuscript text. No AI-generated content was used for research design, data construction, or analysis.

% \section*{Acknowledgments}

\bibliography{custom}

\appendix

\newpage
\section{Filipino Morphological Properties}
\label{sec:filipino-morphology}

We summarize the Filipino properties that motivate PACUTE's task design.

\paragraph{Affixation and non-concatenative processes.}
Filipino (Tagalog-based) word formation is highly productive and relies on a large inventory of affixes, including prefixes (e.g., \textit{mag-}, \textit{nag-}, \textit{pag-}), suffixes (e.g., \textit{-an}, \textit{-in}), circumfixes (e.g., \textit{pag-}\ldots\textit{-an}), and infixes (notably \textit{-um-} and \textit{-in-}) \cite{schachter1983tagalog}. Infixes are inserted inside the root rather than concatenated at an edge, typically but not always after the onset of the first syllable: \textit{kain} $\rightarrow$ \textit{k\textbf{um}ain}, \textit{sulat} $\rightarrow$ \textit{s\textbf{um}ulat} / \textit{s\textbf{in}ulat}. This insertion position is sensitive to the phonological shape of the root (e.g., vowel-initial roots tend to surface with an infix realized prefixally), which complicates purely string-based decomposition.

\paragraph{Reduplication as a productive generator of novel forms.}
Filipino uses reduplication to mark grammatical aspect and related meanings \cite{zamar2022filipino}. Partial reduplication commonly copies the initial CV(C) of the root (\textit{luto} $\rightarrow$ \textit{luluto}, \textit{takbo} $\rightarrow$ \textit{tatakbo}), while full reduplication copies the entire root (\textit{araw-araw}). Because these patterns are productive, they generate many surface forms that may be rare or unseen as wholes, even when the root is frequent.

\paragraph{Morphophonemic alternations at affix boundaries.}
Affixation in Filipino often triggers alternations that obscure a simple affix-root boundary \cite{schachter1983tagalog}. A canonical case is nasal assimilation in the \textit{mang-} prefix, where the nasal assimilates to the place of articulation of the following consonant and may induce deletion or substitution (e.g., \textit{mang-} + \textit{bili} $\rightarrow$ \textit{mamili}). More generally, attachment can involve phonologically conditioned changes that make decomposition a structured reasoning problem rather than a simple string-splitting task.

\paragraph{Stress, glottal stop, and diacritic omission.}
Filipino orthography can mark stress and glottal stop using diacritics (acute, grave, circumflex), yielding minimal pairs such as \textit{bása} ``read'' vs.\ \textit{basâ} ``wet'', \textit{súka} ``vomit'' vs.\ \textit{sukà} ``vinegar,'' and \textit{táyo} ``we'' vs.\ \textit{tayô} ``stand'' \cite{yap1967synchronic}. However, these diacritics are rarely written in contemporary digital text \cite{almario2014masinop}. For modeling, this creates a systematic mismatch between (i) lexicographic forms where distinctions are explicit and (ii) naturally occurring forms where distinctions are collapsed. As a result, tasks that require sensitivity to stress and glottal structure can fail either because the model lacks the concept or because the surface form underdetermines the target.

\paragraph{Implications for evaluation design.}
Taken together, infixation, reduplication, alternations, and diacritic omission create a setting where surface-form competence is not sufficient for morphological competence. A model may (i) recognize characters but fail to segment into morphemes, (ii) memorize frequent affixed words without learning productive rules, or (iii) handle concatenative affixes but fail on infixes because the relevant unit is non-contiguous. PACUTE is designed to disentangle these possibilities by pairing (a) character-level controls, (b) targeted affixation and syllabification tasks, and (c) a hierarchical framework that localizes errors to a specific compositional level.

\section{Prompting and Instance Structure}
\label{sec:prompting}

MCQ instances score options by model log-probability under a fixed prompt and require no special output format. GEN instances use a structured XML output format: the model produces a \texttt{<reflection>} block (1--2 sentences of reasoning) followed by an \texttt{<answer>} block containing only the target form. The evaluator extracts the content inside \texttt{<answer>} and discards the reflection before computing exact-match and contains-match scores. For instruction-tuned models, the format instruction is injected as the system prompt via the tokenizer's chat template; for base models, it is prepended as plain text. Thinking-mode models additionally produce a \texttt{<think>} block prior to the reflection, which is similarly stripped.

The reflection requirement serves two purposes: (i) it provides a lightweight, always-on reasoning trace enabling the qualitative error analysis in \S\ref{sec:error-analysis}; and (ii) it encourages the model to commit to an explicit rationale before producing its answer, which is informative for diagnosing Type III errors (reasoning inconsistency). Prompts are intentionally simple and uniform---no few-shot examples, no chain-of-thought elicitation beyond the reflection format---to preserve comparability across base and instruction-tuned models.

\clearpage
\onecolumn
\section{Model List}
\label{sec:model-list}

Tables~\ref{tab:models-pt} to ~\ref{tab:models-commercial} list all models evaluated in this work with parameter counts, variant types, and references. IT = instruction-tuned; PT = pretrained (base); Thinking = extended chain-of-thought mode.

\begin{table*}[ht]
  \centering
  \caption{Base pre-trained models. Model names link to HuggingFace model pages.}
  \label{tab:models-pt}
  \small
  \begin{tabular}{llrl}
    \toprule
    \textbf{Family} & \textbf{Model}                                                                           & \textbf{Params} & \textbf{Reference}                           \\
    \midrule
    \multirow{4}{*}{Gemma-4}
                    & \href{https://huggingface.co/google/gemma-4-26B-A4B}{google/gemma-4-26B-A4B}             & 26B             & \multirow{4}{*}{\citet{google2025gemma3}}    \\
                    & \href{https://huggingface.co/google/gemma-4-31B}{google/gemma-4-31B}                     & 31B             &                                              \\
                    & \href{https://huggingface.co/google/gemma-4-E2B}{google/gemma-4-E2B}                     & 5B              &                                              \\
                    & \href{https://huggingface.co/google/gemma-4-E4B}{google/gemma-4-E4B}                           & 8B              &                                              \\
    \midrule
    \multirow{4}{*}{GPT-2}
                    & \href{https://huggingface.co/openai-community/gpt2}{openai-community/gpt2}               & 124M            & \multirow{4}{*}{\citet{radford2019language}} \\
                    & \href{https://huggingface.co/openai-community/gpt2-medium}{openai-community/gpt2-medium} & 355M            &                                              \\
                    & \href{https://huggingface.co/openai-community/gpt2-large}{openai-community/gpt2-large}   & 774M            &                                              \\
                    & \href{https://huggingface.co/openai-community/gpt2-xl}{openai-community/gpt2-xl}         & 1.5B            &                                              \\
    \midrule
    \multirow{4}{*}{Qwen-3.5}
                    & \href{https://huggingface.co/Qwen/Qwen3.5-0.8B-Base}{Qwen/Qwen3.5-0.8B-Base}             & 0.8B            & \multirow{4}{*}{\citet{qwen2026qwen35}}      \\
                    & \href{https://huggingface.co/Qwen/Qwen3.5-2B-Base}{Qwen/Qwen3.5-2B-Base}                 & 2B              &                                              \\
                    & \href{https://huggingface.co/Qwen/Qwen3.5-4B-Base}{Qwen/Qwen3.5-4B-Base}            & 4B              &                                              \\
                    & \href{https://huggingface.co/Qwen/Qwen3.5-9B-Base}{Qwen/Qwen3.5-9B-Base}                 & 9B              &                                              \\
    \bottomrule
  \end{tabular}
\end{table*}

\begin{table*}[ht]
  \centering
  \caption{Instruction-tuned and reasoning variants. \textbf{IT} = standard instruction-tuned only; \textbf{IT + Thinking} = evaluated in both standard and thinking modes; \textbf{Thinking} = thinking-only model. Model names link to HuggingFace model pages.}
  \label{tab:models-it}
  \small
  \begin{tabular}{lllrl}
    \toprule
    \textbf{Family} & \textbf{Model}                                                                                           & \textbf{Type} & \textbf{Params} & \textbf{Reference}                                            \\
    \midrule
    \multirow{3}{*}{Gemma-3}
                    & \href{https://huggingface.co/google/gemma-3-4b-it}{google/gemma-3-4b-it}                                 & IT            & 4B              & \multirow{3}{*}{\citet{google2025gemma3}}                     \\
                    & \href{https://huggingface.co/google/gemma-3-12b-it}{google/gemma-3-12b-it}                               & IT            & 12B             &                                                               \\
                    & \href{https://huggingface.co/google/gemma-3-27b-it}{google/gemma-3-27b-it}                               & IT            & 27B             &                                                               \\
    \midrule
    \multirow{4}{*}{Gemma-4}
                    & \href{https://huggingface.co/google/gemma-4-26B-A4B-it}{google/gemma-4-26B-A4B-it}                       & IT + Thinking & 26B             & \multirow{4}{*}{\citet{google2026gemma4}}                     \\
                    & \href{https://huggingface.co/google/gemma-4-31B-it}{google/gemma-4-31B-it}                               & IT + Thinking & 31B             &                                                               \\
                    & \href{https://huggingface.co/google/gemma-4-E2B-it}{google/gemma-4-E2B-it}                               & IT + Thinking & 5B              &                                                               \\
                    & \href{https://huggingface.co/google/gemma-4-E4B-it}{google/gemma-4-E4B-it}                        & IT + Thinking & 8B              &                                                               \\
    \midrule
    \multirow{2}{*}{GPT-OSS}
                    & \href{https://huggingface.co/openai/gpt-oss-20b}{openai/gpt-oss-20b}                                     & Thinking      & 20B             & \multirow{2}{*}{\citet{openai2025gptoss120bgptoss20bmodel}}   \\
                    & \href{https://huggingface.co/openai/gpt-oss-120b}{openai/gpt-oss-120b}                                   & Thinking      & 120B            &                                                               \\
    \midrule
    Mistral-Small-4
                    & \href{https://huggingface.co/mistralai/Mistral-Small-4-119B-2603}{mistralai/Mistral-Small-4-119B-2603}   & IT + Thinking & 119B            & \citet{mistral2026mistralsmall}                               \\
    \midrule
    \multirow{2}{*}{Phi-4}
                    & \href{https://huggingface.co/microsoft/Phi-4-mini-instruct}{microsoft/Phi-4-mini-instruct}               & IT            & 3.8B            & \citet{microsoft2025phi4minitechnicalreportcompact}           \\
                    & \href{https://huggingface.co/microsoft/phi-4}{microsoft/Phi-4}                                           & IT            & 14B             & \citet{abdin2024phi4technicalreport}                          \\
    \midrule
    \multirow{4}{*}{Qwen-3}
                    & \href{https://huggingface.co/Qwen/Qwen3-4B}{Qwen/Qwen3-4B}                                               & IT + Thinking & 4B              & \multirow{4}{*}{\citet{yang2025qwen3technicalreport}}         \\
                    & \href{https://huggingface.co/Qwen/Qwen3-8B}{Qwen/Qwen3-8B}                                               & IT + Thinking & 8B              &                                                               \\
                    & \href{https://huggingface.co/Qwen/Qwen3-14B}{Qwen/Qwen3-14B}                                             & IT + Thinking & 14B             &                                                               \\
                    & \href{https://huggingface.co/Qwen/Qwen3-32B}{Qwen/Qwen3-32B}                                             & IT + Thinking & 32B             &                                                               \\
    \midrule
    Qwen-3.6
                    & \href{https://huggingface.co/Qwen/Qwen3.6-27B}{Qwen/Qwen3.6-27B}                                         & IT + Thinking & 0.6B            & \citet{qwen2026qwen36}                                        \\
    \midrule
    \multirow{2}{*}{SEA-LION-4}
                    & \href{https://huggingface.co/aisingapore/Gemma-SEA-LION-v4.5-W2B-IT}{aisingapore/Gemma-SEA-LION-v4.5-E2B-IT} & IT + Thinking & 5B             & \multirow{2}{*}{\citet{sealion2026sealion45}} \\
                    & \href{https://huggingface.co/aisingapore/Qwen-SEA-LION-v4.5-27B-IT}{aisingapore/Qwen-SEA-LION-v4.5-27B-IT}   & IT + Thinking & 27B             &                                                               \\
    \bottomrule
  \end{tabular}
\end{table*}

\begin{table*}[ht]
  \centering
  \caption{Commercial frontier models.}
  \label{tab:models-commercial}
  \small
  \begin{tabular}{llrl}
    \toprule
    \textbf{Provider} & \textbf{Model}        & \textbf{Params} & \textbf{Reference}                  \\
    \midrule
    \multirow{3}{*}{Anthropic}         & Opus 4.6            & -               & \citet{anthropic2026claudeopus46} \\
    & Sonnet 4.6            & -               & \citet{anthropic2026claudesonnet46} \\
    & Haiku 4.5            & -               & \citet{anthropic2026claudehaiku45} \\
    \midrule
    \multirow{2}{*}{DeepSeek}
                      & DeepSeek R1           & 685B            & \citet{deepseekai2025deepseekr1}    \\
                      & DeepSeek 3.2          & 685B            & \citet{deepseekai2025deepseekv32}   \\
    \midrule
    \multirow{2}{*}{Google}
                      & Gemini-3.5-Flash      & -               & \citet{google2026gemini35flash}     \\
                      & Gemini-3.1-Flash-Lite & -               & \citet{google2026gemini31flashlite} \\
    
    \midrule
    MoonshootAI       & Kimi K2               & 1.1T               & \citet{moonshot2025kimik2}          \\
    \midrule
    \multirow{4}{*}{OpenAI}
                      & GPT-5.5               & -               & \citet{openai2026gpt55}             \\
                      & GPT-5.4-Mini          & -               & \citet{openai2026gpt54mini}         \\
                      & GPT-5.4-Nano          & -               & \citet{openai2026gpt54nano}         \\
                      & GPT-5-Mini            & -               & \citet{openai2025gpt5mini}          \\
    \bottomrule
  \end{tabular}
\end{table*}

\clearpage
\section{Evaluation Results}
\label{sec:evaluation_results}

% --- MCQ table ---
\begin{table*}[ht]
  \centering
  \small
  \setlength{\tabcolsep}{6pt}
  \caption{%
    Human baseline performance and inter-annotator agreement on the
    \textbf{MCQ} format of PACUTE (three annotators).
    \textbf{Mean Acc} is accuracy averaged across annotators ($\pm$std).
    \textbf{Raw Agree} is the fraction of items on which all three annotators
    selected the same option.
    \textbf{Fleiss~$\kappa$} is computed over the four-class option labels
    ($\kappa\!\geq\!0.81$ almost perfect; $0.61$--$0.80$ substantial).
    \textbf{MCQ Avg} is the macro-average across all five benchmarks.%
  }
  \label{tab:human-baselines-mcq}
  \begin{tabular}{lrrr}
    \toprule
    \textbf{Benchmark} & \textbf{Mean Acc} & \textbf{Raw Agree} & \textbf{Fleiss~$\kappa$} \\
    \midrule
    Composition        & $0.926\pm0.03$    & $90.5\%$           & $0.915$                  \\
    Manipulation       & $0.983\pm0.01$    & $98.8\%$           & $0.989$                  \\
    M.\ Extraction     & $0.942\pm0.07$    & $82.5\%$           & $0.842$                  \\
    M.\ Production     & $0.978\pm0.03$    & $93.3\%$           & $0.938$                  \\
    Syllabification    & $0.767\pm0.14$    & $60.0\%$           & $0.589$                  \\
    \midrule
    \textbf{MCQ Avg.}  & $\mathbf{ 0.919}$ & $\mathbf{85.0\%}$  & $\mathbf{0.855}$         \\
    \bottomrule
  \end{tabular}
\end{table*}

% --- GEN table ---
\begin{table*}[ht]
  \centering
  \small
  \setlength{\tabcolsep}{6pt}
  \caption{%
    Human baseline performance and inter-annotator agreement on the
    \textbf{GEN} format of PACUTE (three annotators).
    \textbf{Mean EM} is exact-match accuracy averaged across annotators ($\pm$std).
    \textbf{Raw Agree} is the fraction of items on which all three annotators
    produced identical normalized strings.
    \textbf{Fleiss~$\kappa$} is computed on the binarized correct/incorrect
    judgment against the reference answer
    ($\kappa\!\geq\!0.81$ almost perfect; $0.61$--$0.80$ substantial).
    \textbf{GEN Avg} is the macro-average across all five benchmarks.%
  }
  \label{tab:human-baselines-gen}
  \begin{tabular}{lrrr}
    \toprule
    \textbf{Benchmark} & \textbf{Mean EM} & \textbf{Raw Agree} & \textbf{Fleiss~$\kappa$} \\
    \midrule
    Composition        & $0.982\pm0.00$   & $96.4\%$           & $1.000$                  \\
    Manipulation       & $1.000\pm0.00$   & $100.0\%$          & $1.000$                  \\
    M.\ Extraction     & $0.925\pm0.04$   & $75.0\%$           & $0.765$                  \\
    M.\ Production     & $0.911\pm0.05$   & $86.7\%$           & $0.451$                  \\
    Syllabification    & $0.967\pm0.02$   & $95.0\%$           & $0.483$                  \\
    \midrule
    \textbf{GEN Avg.}  & $\mathbf{0.957}$ & $\mathbf{90.4\%}$  & $\mathbf{0.740}$         \\
    \bottomrule
  \end{tabular}
\end{table*}

\begin{table*}[htbp]
  \centering
  \caption{Results for PT models on PACUTE benchmarks. All values are percentages.}
  \label{tab:results-pt}
  \begin{threeparttable}
    \begin{adjustbox}{max width=\textwidth}
      \begin{tabular}{lrrrrrrrrrrrrrrrr}
        \toprule
        \textbf{Model}               & \multicolumn{2}{c}{\textbf{Comp-MCQ}} & \multicolumn{2}{c}{\textbf{Manip-MCQ}} & \multicolumn{2}{c}{\textbf{MExt-MCQ}} & \multicolumn{2}{c}{\textbf{MProd-MCQ}} & \multicolumn{2}{c}{\textbf{Syll-MCQ}} & \multicolumn{2}{c}{\textbf{Hier-MCQ}} & \multicolumn{2}{c}{\textbf{LGame-MCQ}} & \multicolumn{2}{c}{\textbf{MDA-MCQ}}                                                                                                                                                                                                                                                                                                                                  \\
        \cmidrule(lr){2-3}\cmidrule(lr){4-5}\cmidrule(lr){6-7}\cmidrule(lr){8-9}\cmidrule(lr){10-11}\cmidrule(lr){12-13}\cmidrule(lr){14-15}\cmidrule(lr){16-17}
                                     & \textit{NormAcc}                      & \textit{F1}                            & \textit{NormAcc}                      & \textit{F1}                            & \textit{NormAcc}                      & \textit{F1}                           & \textit{NormAcc}                       & \textit{F1}                           & \textit{NormAcc}                       & \textit{F1}                           & \textit{NormAcc}                     & \textit{F1}                           & \textit{NormAcc}                      & \textit{F1}                           & \textit{NormAcc}                      & \textit{F1}                           \\
        \midrule
        Qwen/Qwen3.5-0.8B-Base       & \cellcolor[HTML]{FDF1EC}16.8          & \cellcolor[HTML]{FDF5F2}54.6           & \cellcolor[HTML]{FAD8C9}2.2           & \cellcolor[HTML]{FBE0D4}42.1           & \cellcolor[HTML]{F4A785}39.0          & \cellcolor[HTML]{F4A785}70.3          & \cellcolor[HTML]{FBE4DA}61.8           & \cellcolor[HTML]{FCECE5}83.3          & \cellcolor[HTML]{92C5DE}\textbf{-45.0} & \cellcolor[HTML]{92C5DE}\textbf{43.1} & \cellcolor[HTML]{ECF5F9}0.2          & \cellcolor[HTML]{EAF3F8}40.2          & \cellcolor[HTML]{F9D2C1}19.2          & \cellcolor[HTML]{FBDED1}56.5          & \cellcolor[HTML]{AAD2E5}51.9          & \cellcolor[HTML]{A4CEE3}78.0          \\
        Qwen/Qwen3.5-2B-Base         & \cellcolor[HTML]{FCE6DC}15.6          & \cellcolor[HTML]{FCEAE2}53.7           & \cellcolor[HTML]{FCEAE2}6.0           & \cellcolor[HTML]{FDF3EF}45.6           & \cellcolor[HTML]{FDEFE9}49.0          & \cellcolor[HTML]{FDF4F0}76.4          & \cellcolor[HTML]{D2E7F1}77.8           & \cellcolor[HTML]{C9E2EE}90.9          & \cellcolor[HTML]{EFF6FA}-48.0          & \cellcolor[HTML]{EDF5F9}41.3          & \cellcolor[HTML]{F4AB8A}-3.3         & \cellcolor[HTML]{F4AB8A}36.7          & \cellcolor[HTML]{FBE3D9}25.5          & \cellcolor[HTML]{FDF1EC}61.2          & \cellcolor[HTML]{A7D0E4}52.8          & \cellcolor[HTML]{A1CDE2}78.5          \\
        Qwen/Qwen3.5-4B-Base         & \cellcolor[HTML]{FDF5F1}17.2          & \cellcolor[HTML]{FEF9F7}55.0           & \cellcolor[HTML]{FEFBFA}9.5           & \cellcolor[HTML]{F7FBFC}48.6           & \cellcolor[HTML]{AFD4E6}60.3          & \cellcolor[HTML]{ACD2E5}82.5          & \cellcolor[HTML]{ABD2E5}85.8           & \cellcolor[HTML]{A7D0E4}94.4          & \cellcolor[HTML]{D0E6F0}-47.0          & \cellcolor[HTML]{CFE5F0}41.9          & \cellcolor[HTML]{F9D0BE}-2.0         & \cellcolor[HTML]{F9D1C0}38.1          & \cellcolor[HTML]{C7E1EE}52.0          & \cellcolor[HTML]{BBDAEA}78.0          & \cellcolor[HTML]{96C7DF}58.0          & \cellcolor[HTML]{95C6DF}81.3          \\
        Qwen/Qwen3.5-9B-Base         & \cellcolor[HTML]{B7D9E9}24.2          & \cellcolor[HTML]{B4D7E8}60.3           & \cellcolor[HTML]{C7E1EE}19.8          & \cellcolor[HTML]{BFDDEB}57.0           & \cellcolor[HTML]{FAFCFD}51.7          & \cellcolor[HTML]{F4F9FB}77.9          & \cellcolor[HTML]{B4D7E8}84.0           & \cellcolor[HTML]{AED4E6}93.6          & \cellcolor[HTML]{FAD8C9}-50.0          & \cellcolor[HTML]{FAD9CA}40.0          & \cellcolor[HTML]{F4A582}-3.6         & \cellcolor[HTML]{F4A582}36.5          & \cellcolor[HTML]{92C5DE}\textbf{68.0} & \cellcolor[HTML]{92C5DE}\textbf{86.4} & \cellcolor[HTML]{92C5DE}\textbf{59.5} & \cellcolor[HTML]{92C5DE}\textbf{82.1} \\
        google/gemma-4-26B-A4B       & \cellcolor[HTML]{FDF1EC}16.8          & \cellcolor[HTML]{FDF5F2}54.6           & \cellcolor[HTML]{92C5DE}\textbf{29.0} & \cellcolor[HTML]{92C5DE}\textbf{63.7}  & \cellcolor[HTML]{A9D1E5}61.0          & \cellcolor[HTML]{A6D0E4}82.9          & \cellcolor[HTML]{9AC9E0}89.3           & \cellcolor[HTML]{99C8E0}95.8          & \cellcolor[HTML]{FAD8C9}-50.0          & \cellcolor[HTML]{FAD9CA}40.0          & \cellcolor[HTML]{FDEFE9}-0.9         & \cellcolor[HTML]{FDF1EB}39.1          & \cellcolor[HTML]{BBDBEA}55.6          & \cellcolor[HTML]{B1D5E7}80.0          & \cellcolor[HTML]{96C7DF}58.0          & \cellcolor[HTML]{95C6DF}81.3          \\
        google/gemma-4-31B           & \cellcolor[HTML]{92C5DE}\textbf{27.4} & \cellcolor[HTML]{92C5DE}\textbf{62.6}  & \cellcolor[HTML]{ADD3E6}24.3          & \cellcolor[HTML]{A8D0E4}60.4           & \cellcolor[HTML]{92C5DE}\textbf{63.7} & \cellcolor[HTML]{92C5DE}\textbf{84.2} & \cellcolor[HTML]{92C5DE}\textbf{91.1}  & \cellcolor[HTML]{92C5DE}\textbf{96.6} & \cellcolor[HTML]{EFF6FA}-48.0          & \cellcolor[HTML]{EDF5F9}41.3          & \cellcolor[HTML]{F7C4AD}-2.4         & \cellcolor[HTML]{F7C5AE}37.6          & \cellcolor[HTML]{9DCBE1}64.5          & \cellcolor[HTML]{9AC9E0}84.7          & \cellcolor[HTML]{A3CEE3}54.0          & \cellcolor[HTML]{9ECBE1}79.2          \\
        google/gemma-4-E2B           & \cellcolor[HTML]{FBDFD3}14.9          & \cellcolor[HTML]{FBE3D9}53.2           & \cellcolor[HTML]{FBDED2}3.5           & \cellcolor[HTML]{FCE7DD}43.3           & \cellcolor[HTML]{FEFDFD}51.0          & \cellcolor[HTML]{FAFCFD}77.5          & \cellcolor[HTML]{D2E7F1}77.8           & \cellcolor[HTML]{C9E2EE}90.9          & \cellcolor[HTML]{F4A582}-52.0          & \cellcolor[HTML]{F4A582}38.7          & \cellcolor[HTML]{F3F9FB}0.0          & \cellcolor[HTML]{F1F7FA}40.0          & \cellcolor[HTML]{FADACC}22.1          & \cellcolor[HTML]{FCE7DE}58.8          & \cellcolor[HTML]{AAD1E5}52.0          & \cellcolor[HTML]{A3CEE3}78.0          \\
        google/gemma-4-E4B           & \cellcolor[HTML]{FDF1EC}16.8          & \cellcolor[HTML]{FDF5F2}54.6           & \cellcolor[HTML]{DFEEF5}15.7          & \cellcolor[HTML]{D5E8F2}53.7           & \cellcolor[HTML]{CFE5F0}56.7          & \cellcolor[HTML]{CAE2EE}80.6          & \cellcolor[HTML]{A3CEE3}87.6           & \cellcolor[HTML]{A0CCE2}95.1          & \cellcolor[HTML]{F4A582}-52.0          & \cellcolor[HTML]{F4A582}38.7          & \cellcolor[HTML]{FDF5F2}-0.7         & \cellcolor[HTML]{FEF7F4}39.4          & \cellcolor[HTML]{F8FBFD}37.2          & \cellcolor[HTML]{E7F2F7}69.2          & \cellcolor[HTML]{B4D7E8}48.8          & \cellcolor[HTML]{ABD2E5}76.2          \\
        openai-community/gpt2        & \cellcolor[HTML]{F4A582}8.9           & \cellcolor[HTML]{F4A582}48.1           & \cellcolor[HTML]{F4A582}-8.5          & \cellcolor[HTML]{F4A582}31.4           & \cellcolor[HTML]{F4A582}38.7          & \cellcolor[HTML]{F4A582}70.1          & \cellcolor[HTML]{F4A582}45.8           & \cellcolor[HTML]{F4A582}74.5          & \cellcolor[HTML]{F7BEA5}-51.0          & \cellcolor[HTML]{F7BFA6}39.4          & \cellcolor[HTML]{B7D9E9}1.8          & \cellcolor[HTML]{B6D8E9}41.7          & \cellcolor[HTML]{F4AC8C}5.5           & \cellcolor[HTML]{F5AF90}45.1          & \cellcolor[HTML]{F6BAA0}0.7           & \cellcolor[HTML]{F7C2AB}40.6          \\
        openai-community/gpt2-large  & \cellcolor[HTML]{F6B99E}11.0          & \cellcolor[HTML]{F6BBA1}49.9           & \cellcolor[HTML]{F6BAA0}-4.0          & \cellcolor[HTML]{F7BEA6}36.1           & \cellcolor[HTML]{F9D4C4}45.3          & \cellcolor[HTML]{FAD9CA}74.2          & \cellcolor[HTML]{F7C4AE}53.8           & \cellcolor[HTML]{F8CAB5}79.0          & \cellcolor[HTML]{F4A582}-52.0          & \cellcolor[HTML]{F4A582}38.7          & \cellcolor[HTML]{99C8E0}2.7          & \cellcolor[HTML]{99C8E0}42.5          & \cellcolor[HTML]{F4A582}2.7           & \cellcolor[HTML]{F4A582}42.5          & \cellcolor[HTML]{F4A582}-7.5          & \cellcolor[HTML]{F4A582}32.5          \\
        openai-community/gpt2-medium & \cellcolor[HTML]{FADBCD}14.5          & \cellcolor[HTML]{FBDFD3}52.8           & \cellcolor[HTML]{F6B69A}-4.8          & \cellcolor[HTML]{F6BA9F}35.2           & \cellcolor[HTML]{F8CBB7}44.0          & \cellcolor[HTML]{F9CEBC}73.4          & \cellcolor[HTML]{F4AC8B}47.6           & \cellcolor[HTML]{F5AD8D}75.5          & \cellcolor[HTML]{F4A582}-52.0          & \cellcolor[HTML]{F4A582}38.7          & \cellcolor[HTML]{92C5DE}\textbf{2.9} & \cellcolor[HTML]{92C5DE}\textbf{42.7} & \cellcolor[HTML]{F5AD8E}5.9           & \cellcolor[HTML]{F5B092}45.4          & \cellcolor[HTML]{F5B497}-1.6          & \cellcolor[HTML]{F6BAA0}38.4          \\
        openai-community/gpt2-xl     & \cellcolor[HTML]{F5AE8F}9.9           & \cellcolor[HTML]{F5AF90}49.0           & \cellcolor[HTML]{F6B599}-5.0          & \cellcolor[HTML]{F6B99E}35.1           & \cellcolor[HTML]{F8C6B0}43.3          & \cellcolor[HTML]{F8C9B5}73.0          & \cellcolor[HTML]{F7BDA4}52.0           & \cellcolor[HTML]{F7C2AA}78.0          & \cellcolor[HTML]{FDF2ED}-49.0          & \cellcolor[HTML]{FDF3EE}40.6          & \cellcolor[HTML]{A8D1E4}2.2          & \cellcolor[HTML]{A7D0E4}42.1          & \cellcolor[HTML]{F4AA89}4.5           & \cellcolor[HTML]{F4AC8B}44.2          & \cellcolor[HTML]{F4A886}-6.1          & \cellcolor[HTML]{F4AA89}33.9          \\
        \bottomrule
      \end{tabular}
    \end{adjustbox}
    \begin{tablenotes}[flushleft]\footnotesize
      \item \textit{Comp} = PACUTE Composition; \textit{Manip} = PACUTE Manipulation; \textit{MExt} = PACUTE Morphological Extraction; \textit{MProd} = 
      \item PACUTE Morphological Production; \textit{Syll} = PACUTE Syllabification; \textit{Hier} = Hierarchical diagnostic; \textit{LGame} = LangGame
      \item (language-agnostic control); \textit{MDA} = Multi-Digit Addition (catastrophic-forgetting probe); Bold = best in column. Cell color:
      \item \colorbox[HTML]{F4A582}{\strut low} $\to$ white $\to$ \colorbox[HTML]{92C5DE}{high} per column.
    \end{tablenotes}
  \end{threeparttable}
\end{table*}

\begin{table*}[htbp]
  \centering
  \caption{Results for IT models on PACUTE MCQ benchmarks. All values are percentages.}
  \label{tab:results-it:mcq}
  \begin{threeparttable}
    \begin{adjustbox}{max width=\textwidth}
      \begin{tabular}{lrrrrrrrrrrrrrrrr}
        \toprule
        \textbf{Model}                                    & \multicolumn{2}{c}{\textbf{Comp-MCQ}} & \multicolumn{2}{c}{\textbf{Manip-MCQ}} & \multicolumn{2}{c}{\textbf{MExt-MCQ}} & \multicolumn{2}{c}{\textbf{MProd-MCQ}} & \multicolumn{2}{c}{\textbf{Syll-MCQ}} & \multicolumn{2}{c}{\textbf{Hier-MCQ}} & \multicolumn{2}{c}{\textbf{LGame-MCQ}} & \multicolumn{2}{c}{\textbf{MDA-MCQ}}                                                                                                                                                                                                                                                                                                                                  \\
        \cmidrule(lr){2-3}\cmidrule(lr){4-5}\cmidrule(lr){6-7}\cmidrule(lr){8-9}\cmidrule(lr){10-11}\cmidrule(lr){12-13}\cmidrule(lr){14-15}\cmidrule(lr){16-17}
                                                          & \textit{NormAcc}                      & \textit{F1}                            & \textit{NormAcc}                      & \textit{F1}                            & \textit{NormAcc}                      & \textit{F1}                           & \textit{NormAcc}                       & \textit{F1}                           & \textit{NormAcc}                       & \textit{F1}                           & \textit{NormAcc}                     & \textit{F1}                           & \textit{NormAcc}                      & \textit{F1}                           & \textit{NormAcc}                      & \textit{F1}                           \\
        \midrule
        Qwen/Qwen3-4B                                     & \cellcolor[HTML]{EAF4F8}18.9          & \cellcolor[HTML]{E1EFF6}56.3           & \cellcolor[HTML]{FCE8DF}8.5           & \cellcolor[HTML]{FDF2ED}47.8           & \cellcolor[HTML]{E9F3F8}48.3          & \cellcolor[HTML]{DCECF4}76.0          & \cellcolor[HTML]{9DCBE1}84.9           & \cellcolor[HTML]{9AC9E0}94.0          & \cellcolor[HTML]{F6B89C}-53.0          & \cellcolor[HTML]{F6BA9F}38.1          & \cellcolor[HTML]{FBE2D7}-3.1         & \cellcolor[HTML]{FBE4DA}37.0          & \cellcolor[HTML]{F3F8FB}47.7          & \cellcolor[HTML]{DBECF4}75.6          & \cellcolor[HTML]{C5E0ED}57.7          & \cellcolor[HTML]{B5D7E8}81.2          \\
        Qwen/Qwen3-4B (thinking)                          & \cellcolor[HTML]{EEF6F9}18.3          & \cellcolor[HTML]{E5F1F7}55.8           & \cellcolor[HTML]{FBE1D6}7.0           & \cellcolor[HTML]{FCEBE4}46.4           & \cellcolor[HTML]{EDF5F9}47.3          & \cellcolor[HTML]{E0EEF5}75.4          & \cellcolor[HTML]{9DCBE1}84.9           & \cellcolor[HTML]{9AC9E0}94.0          & \cellcolor[HTML]{F6B89C}-53.0          & \cellcolor[HTML]{F6BA9F}38.1          & \cellcolor[HTML]{D6E9F2}-0.7         & \cellcolor[HTML]{D4E8F2}39.4          & \cellcolor[HTML]{FEFBF9}40.8          & \cellcolor[HTML]{EBF4F8}71.5          & \cellcolor[HTML]{CEE5F0}54.0          & \cellcolor[HTML]{BCDBEA}79.2          \\
        Qwen/Qwen3-8B                                     & \cellcolor[HTML]{F4F9FB}17.5          & \cellcolor[HTML]{EBF4F8}55.2           & \cellcolor[HTML]{FDF4F0}11.5          & \cellcolor[HTML]{FEFEFE}50.3           & \cellcolor[HTML]{ECF4F9}47.7          & \cellcolor[HTML]{DEEDF5}75.6          & \cellcolor[HTML]{B2D6E7}78.7           & \cellcolor[HTML]{ABD2E5}91.3          & \cellcolor[HTML]{FDF2ED}-44.0          & \cellcolor[HTML]{FEF6F3}43.7          & \cellcolor[HTML]{F4A582}-5.6         & \cellcolor[HTML]{F4A582}34.5          & \cellcolor[HTML]{EBF4F8}51.3          & \cellcolor[HTML]{D4E8F2}77.7          & \cellcolor[HTML]{D6E9F2}50.8          & \cellcolor[HTML]{C2DEEC}77.4          \\
        Qwen/Qwen3-8B (thinking)                          & \cellcolor[HTML]{F4F9FB}17.5          & \cellcolor[HTML]{EBF4F8}55.2           & \cellcolor[HTML]{FDF3EF}11.3          & \cellcolor[HTML]{FEFEFE}50.2           & \cellcolor[HTML]{EEF6FA}47.0          & \cellcolor[HTML]{E1EFF6}75.2          & \cellcolor[HTML]{B2D6E7}78.7           & \cellcolor[HTML]{ABD2E5}91.3          & \cellcolor[HTML]{FCEBE4}-45.0          & \cellcolor[HTML]{FDF0EA}43.1          & \cellcolor[HTML]{F6B599}-4.9         & \cellcolor[HTML]{F6B69A}35.2          & \cellcolor[HTML]{EAF3F8}51.7          & \cellcolor[HTML]{D3E7F1}77.9          & \cellcolor[HTML]{D5E9F2}50.9          & \cellcolor[HTML]{C1DEEC}77.5          \\
        Qwen/Qwen3-14B                                    & \cellcolor[HTML]{F5F9FC}17.3          & \cellcolor[HTML]{ECF4F9}55.1           & \cellcolor[HTML]{F2F8FB}16.5          & \cellcolor[HTML]{E5F1F7}54.4           & \cellcolor[HTML]{D7EAF3}52.7          & \cellcolor[HTML]{CCE4EF}78.4          & \cellcolor[HTML]{9DCBE1}84.9           & \cellcolor[HTML]{9AC9E0}94.0          & \cellcolor[HTML]{FBDED2}-47.0          & \cellcolor[HTML]{FBE3D8}41.9          & \cellcolor[HTML]{FADDD0}-3.3         & \cellcolor[HTML]{FBDED2}36.7          & \cellcolor[HTML]{C5E0ED}67.3          & \cellcolor[HTML]{B6D8E8}86.0          & \cellcolor[HTML]{D1E6F1}52.9          & \cellcolor[HTML]{BEDCEB}78.6          \\
        Qwen/Qwen3-14B (thinking)                         & \cellcolor[HTML]{F5F9FC}17.3          & \cellcolor[HTML]{ECF4F9}55.1           & \cellcolor[HTML]{F4F9FB}16.0          & \cellcolor[HTML]{E8F2F8}54.0           & \cellcolor[HTML]{D7EAF3}52.7          & \cellcolor[HTML]{CCE4EF}78.4          & \cellcolor[HTML]{9DCBE1}84.9           & \cellcolor[HTML]{9AC9E0}94.0          & \cellcolor[HTML]{FBDED2}-47.0          & \cellcolor[HTML]{FBE3D8}41.9          & \cellcolor[HTML]{F7C1A9}-4.4         & \cellcolor[HTML]{F7C2AA}35.6          & \cellcolor[HTML]{C5E0ED}67.6          & \cellcolor[HTML]{B5D7E8}86.2          & \cellcolor[HTML]{CFE5F0}53.6          & \cellcolor[HTML]{BCDBEB}78.9          \\
        Qwen/Qwen3-32B                                    & \cellcolor[HTML]{BDDCEB}25.3          & \cellcolor[HTML]{B8D9E9}61.1           & \cellcolor[HTML]{EAF3F8}18.2          & \cellcolor[HTML]{DDEDF4}55.7           & \cellcolor[HTML]{C5E0ED}57.3          & \cellcolor[HTML]{BBDBEA}81.0          & \cellcolor[HTML]{9AC9E0}85.8           & \cellcolor[HTML]{98C8E0}94.4          & \cellcolor[HTML]{F7BEA5}-52.0          & \cellcolor[HTML]{F7C1A9}38.7          & \cellcolor[HTML]{D6E9F2}-0.7         & \cellcolor[HTML]{D4E8F2}39.4          & \cellcolor[HTML]{C4E0ED}67.7          & \cellcolor[HTML]{B5D7E8}86.2          & \cellcolor[HTML]{BDDCEB}61.2          & \cellcolor[HTML]{AFD4E6}83.0          \\
        Qwen/Qwen3-32B (thinking)                         & \cellcolor[HTML]{BCDBEA}25.5          & \cellcolor[HTML]{B7D8E9}61.2           & \cellcolor[HTML]{EAF3F8}18.2          & \cellcolor[HTML]{DDEDF4}55.7           & \cellcolor[HTML]{C6E0ED}57.0          & \cellcolor[HTML]{BCDBEA}80.8          & \cellcolor[HTML]{9AC9E0}85.8           & \cellcolor[HTML]{98C8E0}94.4          & \cellcolor[HTML]{F7BEA5}-52.0          & \cellcolor[HTML]{F7C1A9}38.7          & \cellcolor[HTML]{DCECF4}-0.9         & \cellcolor[HTML]{DAEBF4}39.1          & \cellcolor[HTML]{C5E0ED}67.5          & \cellcolor[HTML]{B5D8E8}86.1          & \cellcolor[HTML]{BDDCEB}61.1          & \cellcolor[HTML]{AFD4E6}82.9          \\
        Qwen/Qwen3.6-27B                                  & \cellcolor[HTML]{C2DEEC}24.6          & \cellcolor[HTML]{BCDBEA}60.6           & \cellcolor[HTML]{C1DEEC}26.2          & \cellcolor[HTML]{B9D9E9}61.7           & \cellcolor[HTML]{B0D5E7}62.3          & \cellcolor[HTML]{AAD2E5}83.6          & \cellcolor[HTML]{A3CEE3}83.1           & \cellcolor[HTML]{9FCCE2}93.2          & \cellcolor[HTML]{F7C5AE}-51.0          & \cellcolor[HTML]{F8C8B2}39.4          & \cellcolor[HTML]{FBE2D7}-3.1         & \cellcolor[HTML]{FBE4DA}37.0          & \cellcolor[HTML]{9ECBE1}84.1          & \cellcolor[HTML]{9AC9E0}93.7          & \cellcolor[HTML]{CDE4F0}54.3          & \cellcolor[HTML]{BBDBEA}79.3          \\
        Qwen/Qwen3.6-27B (thinking)                       & \cellcolor[HTML]{C4DFED}24.4          & \cellcolor[HTML]{BEDCEB}60.4           & \cellcolor[HTML]{C0DDEC}26.3          & \cellcolor[HTML]{B8D9E9}61.8           & \cellcolor[HTML]{B0D5E7}62.3          & \cellcolor[HTML]{AAD2E5}83.6          & \cellcolor[HTML]{A3CEE3}83.1           & \cellcolor[HTML]{9FCCE2}93.2          & \cellcolor[HTML]{F7C5AE}-51.0          & \cellcolor[HTML]{F8C8B2}39.4          & \cellcolor[HTML]{FBE2D7}-3.1         & \cellcolor[HTML]{FBE4DA}37.0          & \cellcolor[HTML]{9ECBE1}84.1          & \cellcolor[HTML]{9AC9E0}93.7          & \cellcolor[HTML]{CDE4F0}54.3          & \cellcolor[HTML]{BBDBEA}79.3          \\
        aisingapore/Gemma-SEA-LION-v4.5-E2B-IT            & \cellcolor[HTML]{F8CCB8}7.1           & \cellcolor[HTML]{F9D1C0}46.5           & \cellcolor[HTML]{F5B294}-4.5          & \cellcolor[HTML]{F6B598}35.6           & \cellcolor[HTML]{FCECE4}37.3          & \cellcolor[HTML]{FEF7F4}69.3          & \cellcolor[HTML]{F4A988}24.4           & \cellcolor[HTML]{F4AB8B}60.5          & \cellcolor[HTML]{F7BEA5}-52.0          & \cellcolor[HTML]{F7C1A9}38.7          & \cellcolor[HTML]{B4D7E8}0.4          & \cellcolor[HTML]{B2D6E7}40.4          & \cellcolor[HTML]{F6B79B}5.6           & \cellcolor[HTML]{F7C0A8}45.2          & \cellcolor[HTML]{F4AA89}-10.0         & \cellcolor[HTML]{F4AD8D}29.8          \\
        aisingapore/Gemma-SEA-LION-v4.5-E2B-IT (thinking) & \cellcolor[HTML]{F8C5AF}6.0           & \cellcolor[HTML]{F8CAB6}45.5           & \cellcolor[HTML]{F5AE8F}-5.3          & \cellcolor[HTML]{F5B192}34.7           & \cellcolor[HTML]{FDF2EE}39.3          & \cellcolor[HTML]{FEFEFE}70.6          & \cellcolor[HTML]{F4A582}22.7           & \cellcolor[HTML]{F4A582}59.2          & \cellcolor[HTML]{F7BEA5}-52.0          & \cellcolor[HTML]{F7C1A9}38.7          & \cellcolor[HTML]{A6CFE4}0.9          & \cellcolor[HTML]{A5CFE3}40.8          & \cellcolor[HTML]{F6B69A}5.2           & \cellcolor[HTML]{F7BFA6}44.8          & \cellcolor[HTML]{F4AB8A}-9.5          & \cellcolor[HTML]{F5AE8F}30.4          \\
        aisingapore/Qwen-SEA-LION-v4.5-27B-IT             & \cellcolor[HTML]{BBDBEA}25.6          & \cellcolor[HTML]{B6D8E8}61.3           & \cellcolor[HTML]{BBDAEA}27.5          & \cellcolor[HTML]{B3D6E8}62.7           & \cellcolor[HTML]{BDDBEB}59.3          & \cellcolor[HTML]{B4D7E8}82.0          & \cellcolor[HTML]{9AC9E0}85.8           & \cellcolor[HTML]{98C8E0}94.4          & \cellcolor[HTML]{F8CBB7}-50.0          & \cellcolor[HTML]{F9CEBC}40.0          & \cellcolor[HTML]{DCECF4}-0.9         & \cellcolor[HTML]{DAEBF4}39.1          & \cellcolor[HTML]{9BC9E0}85.7          & \cellcolor[HTML]{97C8DF}94.3          & \cellcolor[HTML]{CCE4EF}54.7          & \cellcolor[HTML]{BADAEA}79.5          \\
        aisingapore/Qwen-SEA-LION-v4.5-27B-IT (thinking)  & \cellcolor[HTML]{BBDBEA}25.6          & \cellcolor[HTML]{B6D8E8}61.3           & \cellcolor[HTML]{BCDBEA}27.2          & \cellcolor[HTML]{B5D7E8}62.4           & \cellcolor[HTML]{BEDCEB}59.0          & \cellcolor[HTML]{B5D8E8}81.8          & \cellcolor[HTML]{9AC9E0}85.8           & \cellcolor[HTML]{98C8E0}94.4          & \cellcolor[HTML]{F8CBB7}-50.0          & \cellcolor[HTML]{F9CEBC}40.0          & \cellcolor[HTML]{E3F0F6}-1.1         & \cellcolor[HTML]{E1EFF6}38.9          & \cellcolor[HTML]{9CCAE1}85.2          & \cellcolor[HTML]{98C8E0}94.1          & \cellcolor[HTML]{D0E6F0}53.1          & \cellcolor[HTML]{BDDCEB}78.6          \\
        google/gemma-3-4B-it                              & \cellcolor[HTML]{D6E9F2}21.8          & \cellcolor[HTML]{CEE5F0}58.5           & \cellcolor[HTML]{C5E0ED}25.3          & \cellcolor[HTML]{BDDBEB}61.1           & \cellcolor[HTML]{D9EAF3}52.3          & \cellcolor[HTML]{CDE4F0}78.2          & \cellcolor[HTML]{C4DFED}73.3           & \cellcolor[HTML]{B9DAE9}88.9          & \cellcolor[HTML]{FBE5DB}-46.0          & \cellcolor[HTML]{FCE9E1}42.5          & \cellcolor[HTML]{98C8E0}1.3          & \cellcolor[HTML]{98C8DF}41.3          & \cellcolor[HTML]{E2EFF6}55.1          & \cellcolor[HTML]{CDE4EF}79.7          & \cellcolor[HTML]{C6E0ED}57.3          & \cellcolor[HTML]{B6D8E8}81.0          \\
        google/gemma-3-12B-it                             & \cellcolor[HTML]{D7E9F2}21.7          & \cellcolor[HTML]{CFE5F0}58.4           & \cellcolor[HTML]{9CCAE1}33.5          & \cellcolor[HTML]{9AC9E0}66.8           & \cellcolor[HTML]{B0D5E7}62.3          & \cellcolor[HTML]{AAD2E5}83.6          & \cellcolor[HTML]{9AC9E0}85.8           & \cellcolor[HTML]{98C8E0}94.4          & \cellcolor[HTML]{F6B89C}-53.0          & \cellcolor[HTML]{F6BA9F}38.1          & \cellcolor[HTML]{C1DEEC}0.0          & \cellcolor[HTML]{C0DDEB}40.0          & \cellcolor[HTML]{A8D1E4}79.7          & \cellcolor[HTML]{A1CDE2}91.8          & \cellcolor[HTML]{C3DFEC}58.7          & \cellcolor[HTML]{B3D6E8}81.7          \\
        google/gemma-3-27B-it                             & \cellcolor[HTML]{92C5DE}\textbf{31.6} & \cellcolor[HTML]{92C5DE}\textbf{65.5}  & \cellcolor[HTML]{92C5DE}\textbf{35.7} & \cellcolor[HTML]{92C5DE}\textbf{68.2}  & \cellcolor[HTML]{92C5DE}\textbf{70.0} & \cellcolor[HTML]{92C5DE}\textbf{87.3} & \cellcolor[HTML]{92C5DE}\textbf{88.4}  & \cellcolor[HTML]{92C5DE}\textbf{95.5} & \cellcolor[HTML]{D8EAF3}-37.0          & \cellcolor[HTML]{D3E7F1}47.9          & \cellcolor[HTML]{C1DEEC}0.0          & \cellcolor[HTML]{C0DDEB}40.0          & \cellcolor[HTML]{92C5DE}\textbf{89.6} & \cellcolor[HTML]{92C5DE}\textbf{95.9} & \cellcolor[HTML]{D1E6F1}52.8          & \cellcolor[HTML]{BEDCEB}78.5          \\
        google/gemma-4-26B-A4B-it                         & \cellcolor[HTML]{FAD7C8}9.1           & \cellcolor[HTML]{FADED1}48.2           & \cellcolor[HTML]{F4A582}-7.5          & \cellcolor[HTML]{F4A583}32.5           & \cellcolor[HTML]{FCE9E1}36.7          & \cellcolor[HTML]{FDF5F1}68.9          & \cellcolor[HTML]{F8C7B1}35.1           & \cellcolor[HTML]{F9D0BD}67.8          & \cellcolor[HTML]{F8CBB7}-50.0          & \cellcolor[HTML]{F9CEBC}40.0          & \cellcolor[HTML]{ADD3E6}0.7          & \cellcolor[HTML]{ACD2E5}40.6          & \cellcolor[HTML]{F6B89C}6.0           & \cellcolor[HTML]{F7C1A9}45.6          & \cellcolor[HTML]{F4AB8B}-9.3          & \cellcolor[HTML]{F5AF90}30.5          \\
        google/gemma-4-26B-A4B-it (thinking)              & \cellcolor[HTML]{FAD6C7}8.9           & \cellcolor[HTML]{FADDD0}48.1           & \cellcolor[HTML]{F5AD8D}-5.7          & \cellcolor[HTML]{F5AF90}34.4           & \cellcolor[HTML]{FDF0EA}38.7          & \cellcolor[HTML]{FEFCFA}70.1          & \cellcolor[HTML]{F8CBB8}36.9           & \cellcolor[HTML]{F9D5C5}69.0          & \cellcolor[HTML]{F7BEA5}-52.0          & \cellcolor[HTML]{F7C1A9}38.7          & \cellcolor[HTML]{92C5DE}\textbf{1.6} & \cellcolor[HTML]{92C5DE}\textbf{41.5} & \cellcolor[HTML]{F5B497}3.9           & \cellcolor[HTML]{F6BBA1}43.6          & \cellcolor[HTML]{F5AD8D}-8.4          & \cellcolor[HTML]{F5B193}31.5          \\
        google/gemma-4-31B-it                             & \cellcolor[HTML]{F4A785}0.6           & \cellcolor[HTML]{F4A785}40.6           & \cellcolor[HTML]{F4A582}-7.7          & \cellcolor[HTML]{F4A582}32.3           & \cellcolor[HTML]{FEFAF8}41.7          & \cellcolor[HTML]{F6FAFC}72.0          & \cellcolor[HTML]{FAD8C8}41.3           & \cellcolor[HTML]{FBE3D9}71.8          & \cellcolor[HTML]{E7F2F7}-39.0          & \cellcolor[HTML]{E2EFF6}46.7          & \cellcolor[HTML]{9FCCE2}1.1          & \cellcolor[HTML]{9FCBE1}41.1          & \cellcolor[HTML]{F7C1A9}10.7          & \cellcolor[HTML]{F8CDBA}49.6          & \cellcolor[HTML]{F4A886}-11.1         & \cellcolor[HTML]{F4A988}28.6          \\
        google/gemma-4-31B-it (thinking)                  & \cellcolor[HTML]{F4A582}0.2           & \cellcolor[HTML]{F4A582}40.2           & \cellcolor[HTML]{F4A582}-7.5          & \cellcolor[HTML]{F4A583}32.5           & \cellcolor[HTML]{FEF9F7}41.3          & \cellcolor[HTML]{F7FBFC}71.8          & \cellcolor[HTML]{F9D5C5}40.4           & \cellcolor[HTML]{FBE0D5}71.2          & \cellcolor[HTML]{DFEEF5}-38.0          & \cellcolor[HTML]{DAEBF3}47.3          & \cellcolor[HTML]{98C8E0}1.3          & \cellcolor[HTML]{98C8DF}41.3          & \cellcolor[HTML]{F7C0A8}10.3          & \cellcolor[HTML]{F8CCB9}49.3          & \cellcolor[HTML]{F4A987}-10.5         & \cellcolor[HTML]{F4AB8B}29.2          \\
        google/gemma-4-E2B-it                             & \cellcolor[HTML]{F8CAB6}6.8           & \cellcolor[HTML]{F9D0BD}46.3           & \cellcolor[HTML]{F5B193}-4.7          & \cellcolor[HTML]{F5B497}35.4           & \cellcolor[HTML]{FDF3EF}39.7          & \cellcolor[HTML]{FEFEFE}70.8          & \cellcolor[HTML]{F5B192}27.1           & \cellcolor[HTML]{F5B598}62.4          & \cellcolor[HTML]{F6B89C}-53.0          & \cellcolor[HTML]{F6BA9F}38.1          & \cellcolor[HTML]{C8E2EE}-0.2         & \cellcolor[HTML]{C6E1ED}39.8          & \cellcolor[HTML]{F6B99D}6.4           & \cellcolor[HTML]{F7C2AB}45.9          & \cellcolor[HTML]{F4A988}-10.1         & \cellcolor[HTML]{F4AC8C}29.6          \\
        google/gemma-4-E2B-it (thinking)                  & \cellcolor[HTML]{F8C9B4}6.5           & \cellcolor[HTML]{F9CEBB}46.0           & \cellcolor[HTML]{F5B294}-4.5          & \cellcolor[HTML]{F6B598}35.6           & \cellcolor[HTML]{FDEFE9}38.3          & \cellcolor[HTML]{FEFAF9}69.9          & \cellcolor[HTML]{F5B396}28.0           & \cellcolor[HTML]{F6B89C}63.0          & \cellcolor[HTML]{F6B89C}-53.0          & \cellcolor[HTML]{F6BA9F}38.1          & \cellcolor[HTML]{ADD3E6}0.7          & \cellcolor[HTML]{ACD2E5}40.6          & \cellcolor[HTML]{F6B99E}6.5           & \cellcolor[HTML]{F7C2AB}46.0          & \cellcolor[HTML]{F4A887}-10.7         & \cellcolor[HTML]{F4AB8A}29.1          \\
        google/gemma-4-E4B-it                             & \cellcolor[HTML]{F8C9B5}6.7           & \cellcolor[HTML]{F9CFBC}46.2           & \cellcolor[HTML]{F6BA9F}-2.5          & \cellcolor[HTML]{F7BFA6}37.6           & \cellcolor[HTML]{FAD7C7}31.0          & \cellcolor[HTML]{FBE0D5}65.1          & \cellcolor[HTML]{F9D3C2}39.6           & \cellcolor[HTML]{FADED1}70.7          & \cellcolor[HTML]{FFFFFF}-42.0          & \cellcolor[HTML]{F9FBFD}45.0          & \cellcolor[HTML]{F8FBFC}-1.8         & \cellcolor[HTML]{F5FAFC}38.3          & \cellcolor[HTML]{F4A582}-4.0          & \cellcolor[HTML]{F4A582}36.1          & \cellcolor[HTML]{F4A582}-12.7         & \cellcolor[HTML]{F4A582}26.8          \\
        google/gemma-4-E4B-it (thinking)                  & \cellcolor[HTML]{F8C6B0}6.1           & \cellcolor[HTML]{F8CBB7}45.7           & \cellcolor[HTML]{F6BDA3}-1.8          & \cellcolor[HTML]{F7C2AB}38.2           & \cellcolor[HTML]{FADCCF}32.7          & \cellcolor[HTML]{FCE6DD}66.2          & \cellcolor[HTML]{F9D3C2}39.6           & \cellcolor[HTML]{FADED1}70.7          & \cellcolor[HTML]{FFFFFF}-42.0          & \cellcolor[HTML]{F9FBFD}45.0          & \cellcolor[HTML]{F1F7FA}-1.6         & \cellcolor[HTML]{EFF6FA}38.5          & \cellcolor[HTML]{F4A583}-3.6          & \cellcolor[HTML]{F4A683}36.5          & \cellcolor[HTML]{F4A582}-12.5         & \cellcolor[HTML]{F4A582}27.0          \\
        microsoft/Phi-4-mini-instruct                     & \cellcolor[HTML]{E9F3F8}19.0          & \cellcolor[HTML]{E0EEF5}56.4           & \cellcolor[HTML]{FADDCF}5.8           & \cellcolor[HTML]{FCE6DD}45.4           & \cellcolor[HTML]{FEF8F5}41.0          & \cellcolor[HTML]{F8FBFD}71.6          & \cellcolor[HTML]{B5D7E8}77.8           & \cellcolor[HTML]{ADD3E6}90.9          & \cellcolor[HTML]{FBE5DB}-46.0          & \cellcolor[HTML]{FCE9E1}42.5          & \cellcolor[HTML]{F8FBFC}-1.8         & \cellcolor[HTML]{F5FAFC}38.3          & \cellcolor[HTML]{FDF2ED}36.4          & \cellcolor[HTML]{F5F9FC}68.7          & \cellcolor[HTML]{B7D8E9}63.7          & \cellcolor[HTML]{ABD2E5}84.3          \\
        microsoft/Phi-4                                   & \cellcolor[HTML]{CCE4EF}23.2          & \cellcolor[HTML]{C5E0ED}59.6           & \cellcolor[HTML]{F4F9FB}16.0          & \cellcolor[HTML]{E8F2F8}54.0           & \cellcolor[HTML]{FBE4D9}35.0          & \cellcolor[HTML]{FDEFE9}67.8          & \cellcolor[HTML]{A6CFE4}82.2           & \cellcolor[HTML]{A1CDE2}92.9          & \cellcolor[HTML]{F4A582}-56.0          & \cellcolor[HTML]{F4A582}36.1          & \cellcolor[HTML]{FBE2D7}-3.1         & \cellcolor[HTML]{FBE4DA}37.0          & \cellcolor[HTML]{E3F0F6}54.4          & \cellcolor[HTML]{CEE5F0}79.4          & \cellcolor[HTML]{92C5DE}\textbf{79.6} & \cellcolor[HTML]{92C5DE}\textbf{91.7} \\
        mistralai/Mistral-Small-4-119B-2603               & \cellcolor[HTML]{F8C7B1}6.2           & \cellcolor[HTML]{F8CCB9}45.8           & \cellcolor[HTML]{FBFDFD}14.7          & \cellcolor[HTML]{EEF6FA}52.9           & \cellcolor[HTML]{F5B192}19.7          & \cellcolor[HTML]{F5B497}56.9          & \cellcolor[HTML]{FCEDE7}49.3           & \cellcolor[HTML]{FEFBF9}76.5          & \cellcolor[HTML]{FBE5DB}-46.0          & \cellcolor[HTML]{FCE9E1}42.5          & \cellcolor[HTML]{C8E2EE}-0.2         & \cellcolor[HTML]{C6E1ED}39.8          & \cellcolor[HTML]{F5B295}3.2           & \cellcolor[HTML]{F6B99F}43.0          & \cellcolor[HTML]{F2F8FB}38.8          & \cellcolor[HTML]{DAEBF3}70.2          \\
        mistralai/Mistral-Small-4-119B-2603 (thinking)    & \cellcolor[HTML]{F9D2C1}8.2           & \cellcolor[HTML]{FAD8CA}47.5           & \cellcolor[HTML]{FEF8F6}12.5          & \cellcolor[HTML]{F9FCFD}51.2           & \cellcolor[HTML]{F4A582}16.0          & \cellcolor[HTML]{F4A582}54.0          & \cellcolor[HTML]{FCE6DD}46.7           & \cellcolor[HTML]{FDF3EF}75.0          & \cellcolor[HTML]{92C5DE}\textbf{-28.0} & \cellcolor[HTML]{92C5DE}\textbf{52.9} & \cellcolor[HTML]{FADDD0}-3.3         & \cellcolor[HTML]{FBDED2}36.7          & \cellcolor[HTML]{F4AB8B}-0.4          & \cellcolor[HTML]{F5AF90}39.6          & \cellcolor[HTML]{FEFBF9}31.6          & \cellcolor[HTML]{EAF3F8}65.5          \\
        openai/gpt-oss-120b (thinking)                    & \cellcolor[HTML]{CCE4EF}23.2          & \cellcolor[HTML]{C5E0ED}59.6           & \cellcolor[HTML]{E5F1F7}19.2          & \cellcolor[HTML]{D9EAF3}56.5           & \cellcolor[HTML]{F5B497}20.7          & \cellcolor[HTML]{F6B89D}57.7          & \cellcolor[HTML]{BEDCEB}75.1           & \cellcolor[HTML]{B4D7E8}89.7          & \cellcolor[HTML]{FDF2ED}-44.0          & \cellcolor[HTML]{FEF6F3}43.7          & \cellcolor[HTML]{EAF4F8}-1.3         & \cellcolor[HTML]{E8F2F8}38.7          & \cellcolor[HTML]{E8F2F8}52.5          & \cellcolor[HTML]{D2E7F1}78.3          & \cellcolor[HTML]{97C7DF}77.5          & \cellcolor[HTML]{95C6DE}90.8          \\
        openai/gpt-oss-20b (thinking)                     & \cellcolor[HTML]{CBE3EF}23.4          & \cellcolor[HTML]{C4DFED}59.7           & \cellcolor[HTML]{FDF3EF}11.3          & \cellcolor[HTML]{FEFEFE}50.2           & \cellcolor[HTML]{FCEEE7}38.0          & \cellcolor[HTML]{FEF9F7}69.7          & \cellcolor[HTML]{A9D1E5}81.3           & \cellcolor[HTML]{A3CEE3}92.5          & \cellcolor[HTML]{F4A582}-56.0          & \cellcolor[HTML]{F4A582}36.1          & \cellcolor[HTML]{F8FBFC}-1.8         & \cellcolor[HTML]{F5FAFC}38.3          & \cellcolor[HTML]{FEFCFB}41.3          & \cellcolor[HTML]{E9F3F8}71.8          & \cellcolor[HTML]{9DCBE1}74.8          & \cellcolor[HTML]{99C8E0}89.6          \\
        \bottomrule
      \end{tabular}
    \end{adjustbox}
    \begin{tablenotes}[flushleft]\footnotesize
      \item \textit{Comp} = PACUTE Composition; \textit{Manip} = PACUTE Manipulation; \textit{MExt} = PACUTE Morphological Extraction; \textit{MProd} = 
      \item PACUTE Morphological Production; \textit{Syll} = PACUTE Syllabification; \textit{Hier} = Hierarchical diagnostic; \textit{LGame} = LangGame
      \item (language-agnostic control); \textit{MDA} = Multi-Digit Addition (catastrophic-forgetting probe); Bold = best in column. Cell color:
      \item \colorbox[HTML]{F4A582}{\strut low} $\to$ white $\to$ \colorbox[HTML]{92C5DE}{high} per column.
    \end{tablenotes}
  \end{threeparttable}
\end{table*}

\begin{table*}[htbp]
  \centering
  \caption{Results for IT models on PACUTE GEN benchmarks. EM = exact match; CM = contains match. All values are percentages.}
  \label{tab:results-it:gen}
  \begin{threeparttable}
    \begin{adjustbox}{max width=\textwidth}
      \begin{tabular}{lrrrrrrrrrrrrrrrrrr}
        \toprule
        \textbf{Model}                                    & \multicolumn{2}{c}{\textbf{Comp-Gen}} & \multicolumn{2}{c}{\textbf{Manip-Gen}} & \multicolumn{2}{c}{\textbf{MExt-Gen}} & \multicolumn{2}{c}{\textbf{MProd-Gen}} & \multicolumn{2}{c}{\textbf{Syll-Gen}} & \multicolumn{2}{c}{\textbf{Hier-Gen}} & \multicolumn{2}{c}{\textbf{LGame-Gen}} & \multicolumn{2}{c}{\textbf{MDA-Gen}}  & \multicolumn{2}{c}{\textbf{CUTE-Gen}}                                                                                                                                                                                                                                                                                                                                                                             \\
        \cmidrule(lr){2-3}\cmidrule(lr){4-5}\cmidrule(lr){6-7}\cmidrule(lr){8-9}\cmidrule(lr){10-11}\cmidrule(lr){12-13}\cmidrule(lr){14-15}\cmidrule(lr){16-17}\cmidrule(lr){18-19}
                                                          & \textit{EM}                           & \textit{CM}                            & \textit{EM}                           & \textit{CM}                            & \textit{EM}                           & \textit{CM}                           & \textit{EM}                            & \textit{CM}                           & \textit{EM}                           & \textit{CM}                           & \textit{EM}                           & \textit{CM}                           & \textit{EM}                            & \textit{CM}                            & \textit{EM}                            & \textit{CM}                            & \textit{EM}                           & \textit{CM}                           \\
        \midrule
        Qwen/Qwen3-4B                                     & \cellcolor[HTML]{FDFEFE}70.2          & \cellcolor[HTML]{FBFDFE}70.5           & \cellcolor[HTML]{FDF5F1}47.5          & \cellcolor[HTML]{FEF7F4}49.9           & \cellcolor[HTML]{F8C8B3}33.0          & \cellcolor[HTML]{FCEAE2}48.8          & \cellcolor[HTML]{F9D4C3}19.3           & \cellcolor[HTML]{F9D2C1}19.3          & \cellcolor[HTML]{F4A582}1.5           & \cellcolor[HTML]{C5E0ED}45.0          & \cellcolor[HTML]{E1EFF5}44.7          & \cellcolor[HTML]{DAEBF4}56.8          & \cellcolor[HTML]{A9D1E5}90.6           & \cellcolor[HTML]{A7D0E4}91.6           & \cellcolor[HTML]{92C5DE}99.8           & \cellcolor[HTML]{92C5DE}99.8           & \cellcolor[HTML]{CFE5F0}71.0          & \cellcolor[HTML]{CEE4F0}72.2          \\
        Qwen/Qwen3-4B (thinking)                          & \cellcolor[HTML]{9FCCE2}88.4          & \cellcolor[HTML]{9FCBE1}88.5           & \cellcolor[HTML]{F4A582}9.6           & \cellcolor[HTML]{F4A582}12.0           & \cellcolor[HTML]{FAD7C7}37.8          & \cellcolor[HTML]{FDEFE9}50.2          & \cellcolor[HTML]{FCEBE3}25.3           & \cellcolor[HTML]{FCE9E1}25.3          & \cellcolor[HTML]{FBE5DB}18.5          & \cellcolor[HTML]{F7BFA6}24.0          & \cellcolor[HTML]{B5D7E8}53.0          & \cellcolor[HTML]{B6D8E9}63.8          & \cellcolor[HTML]{F4A582}13.5           & \cellcolor[HTML]{F4A582}14.4           & \cellcolor[HTML]{F4A582}0.8            & \cellcolor[HTML]{F4A582}0.8            & \cellcolor[HTML]{F4A582}7.8           & \cellcolor[HTML]{F4A582}8.3           \\
        Qwen/Qwen3-8B                                     & \cellcolor[HTML]{EFF6FA}72.9          & \cellcolor[HTML]{EFF6FA}72.9           & \cellcolor[HTML]{FEFDFC}51.2          & \cellcolor[HTML]{FEFEFE}52.9           & \cellcolor[HTML]{F9D5C4}37.0          & \cellcolor[HTML]{FDF3EF}51.5          & \cellcolor[HTML]{FEFCFB}30.0           & \cellcolor[HTML]{FEFBF9}30.0          & \cellcolor[HTML]{F6B99E}7.0           & \cellcolor[HTML]{FCE7DD}31.5          & \cellcolor[HTML]{FCE9E0}33.8          & \cellcolor[HTML]{FDEFE8}46.0          & \cellcolor[HTML]{A2CDE2}93.6           & \cellcolor[HTML]{A2CDE2}93.6           & \cellcolor[HTML]{92C5DE}99.9           & \cellcolor[HTML]{92C5DE}99.9           & \cellcolor[HTML]{D3E7F1}69.3          & \cellcolor[HTML]{D3E7F1}69.9          \\
        Qwen/Qwen3-8B (thinking)                          & \cellcolor[HTML]{A1CDE2}88.0          & \cellcolor[HTML]{A1CDE2}88.0           & \cellcolor[HTML]{AED4E6}83.1          & \cellcolor[HTML]{AED3E6}83.8           & \cellcolor[HTML]{FBE4D9}42.0          & \cellcolor[HTML]{FDF5F1}52.0          & \cellcolor[HTML]{FFFFFF}30.7           & \cellcolor[HTML]{FDFEFE}31.3          & \cellcolor[HTML]{EBF4F9}29.5          & \cellcolor[HTML]{FEF7F3}34.5          & \cellcolor[HTML]{BDDCEB}51.5          & \cellcolor[HTML]{BEDCEB}62.3          & \cellcolor[HTML]{92C5DE}99.8           & \cellcolor[HTML]{92C5DE}99.8           & \cellcolor[HTML]{92C5DE}100.0          & \cellcolor[HTML]{92C5DE}100.0          & \cellcolor[HTML]{A3CEE3}88.7          & \cellcolor[HTML]{A4CEE3}89.0          \\
        Qwen/Qwen3-14B                                    & \cellcolor[HTML]{D0E6F0}78.9          & \cellcolor[HTML]{D0E6F0}78.9           & \cellcolor[HTML]{EBF4F9}59.5          & \cellcolor[HTML]{EAF4F8}60.8           & \cellcolor[HTML]{FEFDFC}50.2          & \cellcolor[HTML]{FCFDFE}55.2          & \cellcolor[HTML]{FBFDFE}31.3           & \cellcolor[HTML]{FAFCFD}32.0          & \cellcolor[HTML]{F0F7FA}28.5          & \cellcolor[HTML]{FFFFFF}36.0          & \cellcolor[HTML]{F9FBFD}40.0          & \cellcolor[HTML]{E1EFF6}55.5          & \cellcolor[HTML]{9ECBE1}95.1           & \cellcolor[HTML]{9ECBE1}95.1           & \cellcolor[HTML]{94C6DE}98.7           & \cellcolor[HTML]{94C6DE}98.7           & \cellcolor[HTML]{C5E0ED}75.1          & \cellcolor[HTML]{C5E0ED}75.5          \\
        Qwen/Qwen3-14B (thinking)                         & \cellcolor[HTML]{A2CDE3}87.8          & \cellcolor[HTML]{A2CDE3}87.8           & \cellcolor[HTML]{A9D1E5}85.2          & \cellcolor[HTML]{A8D0E4}85.9           & \cellcolor[HTML]{FBFDFD}51.7          & \cellcolor[HTML]{FCFDFE}55.2          & \cellcolor[HTML]{F2F8FB}33.3           & \cellcolor[HTML]{F4F9FB}33.3          & \cellcolor[HTML]{CDE4F0}36.0          & \cellcolor[HTML]{E8F3F8}39.5          & \cellcolor[HTML]{B9D9E9}52.3          & \cellcolor[HTML]{AAD2E5}66.2          & \cellcolor[HTML]{92C5DE}99.9           & \cellcolor[HTML]{92C5DE}99.9           & \cellcolor[HTML]{92C5DE}\textbf{100.0} & \cellcolor[HTML]{92C5DE}\textbf{100.0} & \cellcolor[HTML]{99C9E0}92.5          & \cellcolor[HTML]{9AC9E0}93.0          \\
        Qwen/Qwen3-32B                                    & \cellcolor[HTML]{C1DEEC}81.8          & \cellcolor[HTML]{C1DEEC}81.8           & \cellcolor[HTML]{E5F1F7}61.9          & \cellcolor[HTML]{E6F1F7}62.5           & \cellcolor[HTML]{FDF4F0}47.2          & \cellcolor[HTML]{FDF3EF}51.5          & \cellcolor[HTML]{FDF5F1}28.0           & \cellcolor[HTML]{FDF3EF}28.0          & \cellcolor[HTML]{FEFCFB}24.5          & \cellcolor[HTML]{EEF6FA}38.5          & \cellcolor[HTML]{E9F3F8}43.0          & \cellcolor[HTML]{DBECF4}56.7          & \cellcolor[HTML]{9DCBE1}95.3           & \cellcolor[HTML]{9DCBE1}95.4           & \cellcolor[HTML]{92C5DE}100.0          & \cellcolor[HTML]{92C5DE}100.0          & \cellcolor[HTML]{BCDBEA}78.6          & \cellcolor[HTML]{BDDBEB}79.1          \\
        Qwen/Qwen3-32B (thinking)                         & \cellcolor[HTML]{A4CEE3}87.5          & \cellcolor[HTML]{A4CEE3}87.5           & \cellcolor[HTML]{ADD3E6}83.9          & \cellcolor[HTML]{ADD3E6}84.1           & \cellcolor[HTML]{F6FAFC}53.0          & \cellcolor[HTML]{F7FBFC}56.5          & \cellcolor[HTML]{FFFFFF}30.7           & \cellcolor[HTML]{FEFDFD}30.7          & \cellcolor[HTML]{E2EFF6}31.5          & \cellcolor[HTML]{D5E8F2}42.5          & \cellcolor[HTML]{AFD4E6}54.2          & \cellcolor[HTML]{A4CFE3}67.3          & \cellcolor[HTML]{95C6DF}98.6           & \cellcolor[HTML]{92C5DE}99.9           & \cellcolor[HTML]{92C5DE}100.0          & \cellcolor[HTML]{92C5DE}100.0          & \cellcolor[HTML]{99C9E0}92.6          & \cellcolor[HTML]{9BC9E0}92.7          \\
        Qwen/Qwen3.6-27B                                  & \cellcolor[HTML]{A3CEE3}87.6          & \cellcolor[HTML]{A3CEE3}87.6           & \cellcolor[HTML]{C3DFEC}75.2          & \cellcolor[HTML]{C1DEEC}76.5           & \cellcolor[HTML]{F6FAFC}53.0          & \cellcolor[HTML]{EAF3F8}59.8          & \cellcolor[HTML]{D7EAF3}39.3           & \cellcolor[HTML]{D5E8F2}40.0          & \cellcolor[HTML]{A8D1E4}44.0          & \cellcolor[HTML]{B8D9E9}47.0          & \cellcolor[HTML]{B1D5E7}53.8          & \cellcolor[HTML]{97C8DF}69.8          & \cellcolor[HTML]{93C5DE}99.5           & \cellcolor[HTML]{92C5DE}99.8           & \cellcolor[HTML]{92C5DE}99.9           & \cellcolor[HTML]{92C5DE}99.9           & \cellcolor[HTML]{B3D6E8}82.1          & \cellcolor[HTML]{AED3E6}85.1          \\
        Qwen/Qwen3.6-27B (thinking)                       & \cellcolor[HTML]{9DCAE1}88.9          & \cellcolor[HTML]{9DCAE1}88.9           & \cellcolor[HTML]{97C7DF}92.4          & \cellcolor[HTML]{96C7DF}92.6           & \cellcolor[HTML]{E4F0F6}58.0          & \cellcolor[HTML]{E5F1F7}60.8          & \cellcolor[HTML]{B6D8E9}46.7           & \cellcolor[HTML]{B6D8E9}46.7          & \cellcolor[HTML]{C9E2EE}37.0          & \cellcolor[HTML]{D5E8F2}42.5          & \cellcolor[HTML]{A9D1E5}55.3          & \cellcolor[HTML]{A3CEE3}67.5          & \cellcolor[HTML]{92C5DE}99.8           & \cellcolor[HTML]{92C5DE}99.8           & \cellcolor[HTML]{92C5DE}100.0          & \cellcolor[HTML]{92C5DE}100.0          & \cellcolor[HTML]{D7EAF3}67.6          & \cellcolor[HTML]{92C5DE}96.1          \\
        aisingapore/Gemma-SEA-LION-v4.5-E2B-IT            & \cellcolor[HTML]{FEFDFC}69.5          & \cellcolor[HTML]{FEFDFC}69.5           & \cellcolor[HTML]{F6FAFC}55.2          & \cellcolor[HTML]{F4F9FB}57.2           & \cellcolor[HTML]{FAD8C8}38.0          & \cellcolor[HTML]{FEF9F7}53.2          & \cellcolor[HTML]{FCEDE6}26.0           & \cellcolor[HTML]{FCEEE8}26.7          & \cellcolor[HTML]{FADACB}15.5          & \cellcolor[HTML]{FDF4F0}34.0          & \cellcolor[HTML]{EEF5F9}42.2          & \cellcolor[HTML]{D9EBF3}57.0          & \cellcolor[HTML]{A2CDE3}93.3           & \cellcolor[HTML]{A0CCE2}94.2           & \cellcolor[HTML]{93C5DE}99.5           & \cellcolor[HTML]{93C5DE}99.5           & \cellcolor[HTML]{D7E9F2}67.9          & \cellcolor[HTML]{D2E7F1}70.4          \\
        aisingapore/Gemma-SEA-LION-v4.5-E2B-IT (thinking) & \cellcolor[HTML]{FEFBF9}68.9          & \cellcolor[HTML]{FEFAF9}68.9           & \cellcolor[HTML]{F5FAFC}55.5          & \cellcolor[HTML]{F2F8FB}57.8           & \cellcolor[HTML]{FAD7C7}37.8          & \cellcolor[HTML]{FEF7F4}52.5          & \cellcolor[HTML]{FBE6DC}24.0           & \cellcolor[HTML]{FCE6DD}24.7          & \cellcolor[HTML]{FAD8C9}15.0          & \cellcolor[HTML]{FEFCFB}35.5          & \cellcolor[HTML]{EEF6FA}42.0          & \cellcolor[HTML]{D7E9F2}57.5          & \cellcolor[HTML]{A0CCE2}94.4           & \cellcolor[HTML]{A0CCE2}94.5           & \cellcolor[HTML]{93C5DE}99.4           & \cellcolor[HTML]{93C5DE}99.4           & \cellcolor[HTML]{D9EAF3}67.1          & \cellcolor[HTML]{D3E7F1}70.1          \\
        aisingapore/Qwen-SEA-LION-v4.5-27B-IT             & \cellcolor[HTML]{A0CCE2}88.2          & \cellcolor[HTML]{A0CCE2}88.2           & \cellcolor[HTML]{BCDBEA}77.9          & \cellcolor[HTML]{BCDBEA}78.5           & \cellcolor[HTML]{EDF5F9}55.5          & \cellcolor[HTML]{E8F3F8}60.0          & \cellcolor[HTML]{CBE3EF}42.0           & \cellcolor[HTML]{CCE4EF}42.0          & \cellcolor[HTML]{D9EAF3}33.5          & \cellcolor[HTML]{FBFDFE}36.5          & \cellcolor[HTML]{B9DAEA}52.2          & \cellcolor[HTML]{A3CEE3}67.5          & \cellcolor[HTML]{93C5DE}99.3           & \cellcolor[HTML]{93C5DE}99.5           & \cellcolor[HTML]{92C5DE}100.0          & \cellcolor[HTML]{92C5DE}100.0          & \cellcolor[HTML]{B0D5E7}83.4          & \cellcolor[HTML]{ACD3E6}85.8          \\
        aisingapore/Qwen-SEA-LION-v4.5-27B-IT (thinking)  & \cellcolor[HTML]{9DCAE1}88.9          & \cellcolor[HTML]{9DCAE1}88.9           & \cellcolor[HTML]{9DCBE1}89.8          & \cellcolor[HTML]{9ECBE1}89.8           & \cellcolor[HTML]{E5F1F7}57.8          & \cellcolor[HTML]{E4F1F7}61.0          & \cellcolor[HTML]{BFDDEB}44.7           & \cellcolor[HTML]{C0DDEB}44.7          & \cellcolor[HTML]{F6FAFC}27.0          & \cellcolor[HTML]{FAD9CB}29.0          & \cellcolor[HTML]{A2CDE2}56.7          & \cellcolor[HTML]{A1CDE2}68.0          & \cellcolor[HTML]{98C8DF}97.6           & \cellcolor[HTML]{98C8DF}97.6           & \cellcolor[HTML]{95C6DE}98.6           & \cellcolor[HTML]{95C6DE}98.6           & \cellcolor[HTML]{C8E2EE}73.6          & \cellcolor[HTML]{97C7DF}94.3          \\
        google/gemma-3-4B-it                              & \cellcolor[HTML]{F7C0A8}55.1          & \cellcolor[HTML]{F7C0A7}55.1           & \cellcolor[HTML]{FCEEE7}44.0          & \cellcolor[HTML]{FDEFE9}46.1           & \cellcolor[HTML]{FCE7DE}43.0          & \cellcolor[HTML]{FEF9F7}53.2          & \cellcolor[HTML]{F7C3AB}14.7           & \cellcolor[HTML]{F7C0A8}14.7          & \cellcolor[HTML]{FADACB}15.5          & \cellcolor[HTML]{AED4E6}48.5          & \cellcolor[HTML]{F7BEA5}24.0          & \cellcolor[HTML]{F8C9B4}37.0          & \cellcolor[HTML]{ADD3E6}88.9           & \cellcolor[HTML]{AED4E6}88.9           & \cellcolor[HTML]{C3DFED}77.3           & \cellcolor[HTML]{C3DFED}77.4           & \cellcolor[HTML]{FEFEFD}51.4          & \cellcolor[HTML]{F2F8FB}57.3          \\
        google/gemma-3-12B-it                             & \cellcolor[HTML]{FCFDFE}70.4          & \cellcolor[HTML]{FCFDFE}70.4           & \cellcolor[HTML]{F7FBFC}54.9          & \cellcolor[HTML]{F7FBFC}55.9           & \cellcolor[HTML]{FEF7F4}48.2          & \cellcolor[HTML]{F5F9FC}57.0          & \cellcolor[HTML]{E9F3F8}35.3           & \cellcolor[HTML]{EBF4F8}35.3          & \cellcolor[HTML]{F4F9FB}27.5          & \cellcolor[HTML]{B2D6E7}48.0          & \cellcolor[HTML]{FDF6F2}36.8          & \cellcolor[HTML]{FDF2ED}46.8          & \cellcolor[HTML]{99C9E0}96.9           & \cellcolor[HTML]{99C9E0}96.9           & \cellcolor[HTML]{95C6DE}98.5           & \cellcolor[HTML]{95C6DE}98.5           & \cellcolor[HTML]{CBE3EF}72.7          & \cellcolor[HTML]{CAE2EF}73.7          \\
        google/gemma-3-27B-it                             & \cellcolor[HTML]{E0EEF5}75.8          & \cellcolor[HTML]{E0EEF5}75.8           & \cellcolor[HTML]{EBF4F9}59.6          & \cellcolor[HTML]{EBF4F9}60.6           & \cellcolor[HTML]{F2F8FB}54.0          & \cellcolor[HTML]{D8EAF3}64.0          & \cellcolor[HTML]{BFDDEB}44.7           & \cellcolor[HTML]{C0DDEB}44.7          & \cellcolor[HTML]{FDF0EB}21.5          & \cellcolor[HTML]{FBDFD2}30.0          & \cellcolor[HTML]{FEFEFE}39.0          & \cellcolor[HTML]{E7F2F7}54.3          & \cellcolor[HTML]{99C9E0}96.9           & \cellcolor[HTML]{99C9E0}96.9           & \cellcolor[HTML]{92C5DE}100.0          & \cellcolor[HTML]{92C5DE}100.0          & \cellcolor[HTML]{BFDDEB}77.2          & \cellcolor[HTML]{C0DDEC}77.8          \\
        google/gemma-4-26B-A4B-it                         & \cellcolor[HTML]{9FCBE1}88.5          & \cellcolor[HTML]{9FCBE1}88.5           & \cellcolor[HTML]{B3D6E8}81.2          & \cellcolor[HTML]{B4D7E8}81.4           & \cellcolor[HTML]{95C6DF}79.2          & \cellcolor[HTML]{96C7DF}79.8          & \cellcolor[HTML]{9BC9E0}52.7           & \cellcolor[HTML]{9BC9E0}52.7          & \cellcolor[HTML]{FDFEFE}25.5          & \cellcolor[HTML]{F8C7B1}25.5          & \cellcolor[HTML]{B1D5E7}53.8          & \cellcolor[HTML]{92C5DE}\textbf{71.0} & \cellcolor[HTML]{92C5DE}99.7           & \cellcolor[HTML]{92C5DE}\textbf{100.0} & \cellcolor[HTML]{93C6DE}99.1           & \cellcolor[HTML]{93C5DE}99.2           & \cellcolor[HTML]{A0CCE2}89.8          & \cellcolor[HTML]{9ECBE1}91.5          \\
        google/gemma-4-26B-A4B-it (thinking)              & \cellcolor[HTML]{9ECBE1}88.7          & \cellcolor[HTML]{9ECBE1}88.7           & \cellcolor[HTML]{B4D7E8}81.1          & \cellcolor[HTML]{B4D7E8}81.4           & \cellcolor[HTML]{94C6DE}79.5          & \cellcolor[HTML]{94C6DE}80.2          & \cellcolor[HTML]{9BC9E0}52.7           & \cellcolor[HTML]{9BC9E0}52.7          & \cellcolor[HTML]{FEFEFD}25.0          & \cellcolor[HTML]{F7C4AE}25.0          & \cellcolor[HTML]{B3D6E7}53.5          & \cellcolor[HTML]{9BCAE0}69.2          & \cellcolor[HTML]{92C5DE}\textbf{100.0} & \cellcolor[HTML]{92C5DE}100.0          & \cellcolor[HTML]{93C6DE}99.1           & \cellcolor[HTML]{93C6DE}99.1           & \cellcolor[HTML]{9ECBE1}90.8          & \cellcolor[HTML]{9ECBE1}91.3          \\
        google/gemma-4-31B-it                             & \cellcolor[HTML]{A2CDE3}87.8          & \cellcolor[HTML]{A2CDE3}87.8           & \cellcolor[HTML]{A3CEE3}87.5          & \cellcolor[HTML]{A3CEE3}87.6           & \cellcolor[HTML]{92C5DE}80.0          & \cellcolor[HTML]{93C5DE}80.5          & \cellcolor[HTML]{92C5DE}\textbf{54.7}  & \cellcolor[HTML]{92C5DE}\textbf{54.7} & \cellcolor[HTML]{94C6DE}48.5          & \cellcolor[HTML]{A2CDE2}50.5          & \cellcolor[HTML]{B4D7E8}53.2          & \cellcolor[HTML]{98C8E0}69.7          & \cellcolor[HTML]{92C5DE}99.9           & \cellcolor[HTML]{92C5DE}99.9           & \cellcolor[HTML]{92C5DE}100.0          & \cellcolor[HTML]{92C5DE}100.0          & \cellcolor[HTML]{DAEBF3}66.6          & \cellcolor[HTML]{98C8DF}93.9          \\
        google/gemma-4-31B-it (thinking)                  & \cellcolor[HTML]{A2CDE3}87.8          & \cellcolor[HTML]{A2CDE3}87.8           & \cellcolor[HTML]{A4CEE3}87.4          & \cellcolor[HTML]{A4CEE3}87.5           & \cellcolor[HTML]{92C5DE}\textbf{80.2} & \cellcolor[HTML]{92C5DE}\textbf{80.8} & \cellcolor[HTML]{95C6DE}54.0           & \cellcolor[HTML]{95C6DE}54.0          & \cellcolor[HTML]{92C5DE}\textbf{49.0} & \cellcolor[HTML]{98C8DF}52.0          & \cellcolor[HTML]{B2D6E7}53.7          & \cellcolor[HTML]{9AC9E0}69.3          & \cellcolor[HTML]{92C5DE}99.9           & \cellcolor[HTML]{92C5DE}99.9           & \cellcolor[HTML]{92C5DE}100.0          & \cellcolor[HTML]{92C5DE}100.0          & \cellcolor[HTML]{DAEBF3}66.6          & \cellcolor[HTML]{98C8DF}94.1          \\
        google/gemma-4-E2B-it                             & \cellcolor[HTML]{F9FCFD}70.9          & \cellcolor[HTML]{F9FCFD}70.9           & \cellcolor[HTML]{FAFCFD}53.8          & \cellcolor[HTML]{F7FBFC}56.0           & \cellcolor[HTML]{FADED1}40.0          & \cellcolor[HTML]{FDFEFE}55.0          & \cellcolor[HTML]{FDF2ED}27.3           & \cellcolor[HTML]{FDF1EB}27.3          & \cellcolor[HTML]{F9D4C3}14.0          & \cellcolor[HTML]{FDF1EC}33.5          & \cellcolor[HTML]{DAEBF3}46.0          & \cellcolor[HTML]{CBE3EF}59.8          & \cellcolor[HTML]{A2CDE3}93.4           & \cellcolor[HTML]{A0CCE2}94.2           & \cellcolor[HTML]{93C5DE}99.5           & \cellcolor[HTML]{93C5DE}99.5           & \cellcolor[HTML]{D9EAF3}66.9          & \cellcolor[HTML]{D3E7F1}70.1          \\
        google/gemma-4-E2B-it (thinking)                  & \cellcolor[HTML]{FDFEFE}70.2          & \cellcolor[HTML]{FDFEFE}70.2           & \cellcolor[HTML]{FBFDFE}53.2          & \cellcolor[HTML]{F8FBFD}55.5           & \cellcolor[HTML]{FBDED2}40.2          & \cellcolor[HTML]{FCFDFE}55.2          & \cellcolor[HTML]{FDF2ED}27.3           & \cellcolor[HTML]{FDF1EB}27.3          & \cellcolor[HTML]{FBDFD3}17.0          & \cellcolor[HTML]{FCE9E1}32.0          & \cellcolor[HTML]{DFEEF5}45.0          & \cellcolor[HTML]{CEE5F0}59.2          & \cellcolor[HTML]{A1CDE2}93.8           & \cellcolor[HTML]{A1CDE2}93.9           & \cellcolor[HTML]{93C5DE}99.4           & \cellcolor[HTML]{93C5DE}99.4           & \cellcolor[HTML]{D9EAF3}67.1          & \cellcolor[HTML]{D2E7F1}70.4          \\
        google/gemma-4-E4B-it                             & \cellcolor[HTML]{C7E1EE}80.7          & \cellcolor[HTML]{C7E1EE}80.7           & \cellcolor[HTML]{EBF4F9}59.6          & \cellcolor[HTML]{E9F3F8}61.1           & \cellcolor[HTML]{FDF0EA}46.0          & \cellcolor[HTML]{FDF4F0}51.7          & \cellcolor[HTML]{D7EAF3}39.3           & \cellcolor[HTML]{D8EAF3}39.3          & \cellcolor[HTML]{DBECF4}33.0          & \cellcolor[HTML]{FEFCFB}35.5          & \cellcolor[HTML]{CCE3EF}48.7          & \cellcolor[HTML]{B4D7E8}64.3          & \cellcolor[HTML]{9FCCE2}94.5           & \cellcolor[HTML]{A0CCE2}94.5           & \cellcolor[HTML]{92C5DE}99.8           & \cellcolor[HTML]{92C5DE}99.8           & \cellcolor[HTML]{C0DDEC}76.9          & \cellcolor[HTML]{BEDCEB}78.5          \\
        google/gemma-4-E4B-it (thinking)                  & \cellcolor[HTML]{C8E2EE}80.4          & \cellcolor[HTML]{C9E2EE}80.4           & \cellcolor[HTML]{EAF4F8}59.9          & \cellcolor[HTML]{E9F3F8}61.4           & \cellcolor[HTML]{FDEFE9}45.8          & \cellcolor[HTML]{FDF5F1}52.0          & \cellcolor[HTML]{D4E8F2}40.0           & \cellcolor[HTML]{D5E8F2}40.0          & \cellcolor[HTML]{E0EEF5}32.0          & \cellcolor[HTML]{F2F8FB}38.0          & \cellcolor[HTML]{CDE4EF}48.5          & \cellcolor[HTML]{B2D6E7}64.7          & \cellcolor[HTML]{9FCCE2}94.6           & \cellcolor[HTML]{9FCCE2}94.6           & \cellcolor[HTML]{92C5DE}99.8           & \cellcolor[HTML]{92C5DE}99.8           & \cellcolor[HTML]{BFDDEB}77.2          & \cellcolor[HTML]{BFDCEB}78.3          \\
        microsoft/Phi-4-mini-instruct                     & \cellcolor[HTML]{F4A582}48.5          & \cellcolor[HTML]{F4A582}48.7           & \cellcolor[HTML]{FBE0D4}37.5          & \cellcolor[HTML]{FBE1D5}39.6           & \cellcolor[HTML]{F5B091}25.0          & \cellcolor[HTML]{FADACC}44.2          & \cellcolor[HTML]{F6BBA1}12.7           & \cellcolor[HTML]{F7BEA5}14.0          & \cellcolor[HTML]{F6BBA1}7.5           & \cellcolor[HTML]{EEF6FA}38.5          & \cellcolor[HTML]{F4A582}18.0          & \cellcolor[HTML]{F5AE8F}30.8          & \cellcolor[HTML]{D5E8F2}73.3           & \cellcolor[HTML]{D5E9F2}73.3           & \cellcolor[HTML]{92C5DE}99.9           & \cellcolor[HTML]{92C5DE}99.9           & \cellcolor[HTML]{FCEBE3}42.2          & \cellcolor[HTML]{FDF1EC}45.7          \\
        microsoft/phi-4                                   & \cellcolor[HTML]{A9D1E5}86.5          & \cellcolor[HTML]{A9D1E5}86.5           & \cellcolor[HTML]{E0EEF5}64.0          & \cellcolor[HTML]{E1EFF5}64.5           & \cellcolor[HTML]{FCE8DF}43.2          & \cellcolor[HTML]{FDEFE9}50.2          & \cellcolor[HTML]{FBFDFE}31.3           & \cellcolor[HTML]{F7FAFC}32.7          & \cellcolor[HTML]{F6FAFC}27.0          & \cellcolor[HTML]{FFFFFF}36.0          & \cellcolor[HTML]{EEF6FA}42.0          & \cellcolor[HTML]{C1DEEC}61.7          & \cellcolor[HTML]{9DCBE1}95.5           & \cellcolor[HTML]{9DCBE1}95.5           & \cellcolor[HTML]{92C5DE}99.9           & \cellcolor[HTML]{92C5DE}99.9           & \cellcolor[HTML]{C3DFEC}75.9          & \cellcolor[HTML]{C3DFEC}76.5          \\
        mistralai/Mistral-Small-4-119B-2603               & \cellcolor[HTML]{C6E0ED}80.9          & \cellcolor[HTML]{C6E0ED}80.9           & \cellcolor[HTML]{DDEDF4}64.9          & \cellcolor[HTML]{DCECF4}66.1           & \cellcolor[HTML]{E5F1F7}57.8          & \cellcolor[HTML]{DDEDF4}62.7          & \cellcolor[HTML]{D1E6F1}40.7           & \cellcolor[HTML]{CFE5F0}41.3          & \cellcolor[HTML]{C2DEEC}38.5          & \cellcolor[HTML]{A2CDE2}50.5          & \cellcolor[HTML]{D0E6F0}47.8          & \cellcolor[HTML]{CBE3EF}59.8          & \cellcolor[HTML]{9BC9E0}96.3           & \cellcolor[HTML]{9BCAE0}96.3           & \cellcolor[HTML]{93C6DE}99.1           & \cellcolor[HTML]{93C6DE}99.1           & \cellcolor[HTML]{BDDCEB}78.1          & \cellcolor[HTML]{BEDCEB}78.4          \\
        mistralai/Mistral-Small-4-119B-2603 (thinking)    & \cellcolor[HTML]{C1DEEC}81.8          & \cellcolor[HTML]{C1DEEC}81.8           & \cellcolor[HTML]{DCECF4}65.4          & \cellcolor[HTML]{DCECF4}66.1           & \cellcolor[HTML]{E7F2F8}57.0          & \cellcolor[HTML]{DFEEF5}62.3          & \cellcolor[HTML]{CBE3EF}42.0           & \cellcolor[HTML]{C9E2EE}42.7          & \cellcolor[HTML]{CDE4F0}36.0          & \cellcolor[HTML]{92C5DE}\textbf{53.0} & \cellcolor[HTML]{DAEBF3}46.0          & \cellcolor[HTML]{D5E8F2}57.8          & \cellcolor[HTML]{9BCAE0}96.1           & \cellcolor[HTML]{9BCAE1}96.1           & \cellcolor[HTML]{96C7DF}98.0           & \cellcolor[HTML]{96C7DF}98.0           & \cellcolor[HTML]{BADAEA}79.4          & \cellcolor[HTML]{BBDAEA}79.8          \\
        openai/gpt-oss-120b (thinking)                    & \cellcolor[HTML]{92C5DE}\textbf{91.1} & \cellcolor[HTML]{92C5DE}\textbf{91.1}  & \cellcolor[HTML]{92C5DE}\textbf{94.4} & \cellcolor[HTML]{92C5DE}\textbf{94.4}  & \cellcolor[HTML]{CBE3EF}64.8          & \cellcolor[HTML]{CBE3EF}67.0          & \cellcolor[HTML]{98C8DF}53.3           & \cellcolor[HTML]{98C8DF}53.3          & \cellcolor[HTML]{A2CDE2}45.5          & \cellcolor[HTML]{BEDCEB}46.0          & \cellcolor[HTML]{92C5DE}\textbf{59.8} & \cellcolor[HTML]{92C5DE}70.8          & \cellcolor[HTML]{92C5DE}99.7           & \cellcolor[HTML]{92C5DE}99.7           & \cellcolor[HTML]{92C5DE}100.0          & \cellcolor[HTML]{92C5DE}100.0          & \cellcolor[HTML]{92C5DE}\textbf{95.7} & \cellcolor[HTML]{92C5DE}\textbf{96.5} \\
        openai/gpt-oss-20b (thinking)                     & \cellcolor[HTML]{9CCAE1}89.1          & \cellcolor[HTML]{9CCAE1}89.1           & \cellcolor[HTML]{B1D5E7}82.0          & \cellcolor[HTML]{B2D6E7}82.2           & \cellcolor[HTML]{F4A582}21.2          & \cellcolor[HTML]{F4A582}28.7          & \cellcolor[HTML]{F4A582}6.7            & \cellcolor[HTML]{F4A582}7.3           & \cellcolor[HTML]{F7C1A9}9.0           & \cellcolor[HTML]{F4A582}19.0          & \cellcolor[HTML]{F7C0A8}24.5          & \cellcolor[HTML]{F4A582}28.5          & \cellcolor[HTML]{9ECBE1}95.2           & \cellcolor[HTML]{9BCAE0}96.2           & \cellcolor[HTML]{92C5DE}100.0          & \cellcolor[HTML]{92C5DE}100.0          & \cellcolor[HTML]{F7BFA7}20.9          & \cellcolor[HTML]{F7C4AD}23.7          \\
        \bottomrule
      \end{tabular}
    \end{adjustbox}
    \begin{tablenotes}[flushleft]\footnotesize
      \item \textit{Comp} = PACUTE Composition; \textit{Manip} = PACUTE Manipulation; \textit{MExt} = PACUTE Morphological Extraction; \textit{MProd} = 
      \item PACUTE Morphological Production; \textit{Syll} = PACUTE Syllabification; \textit{Hier} = Hierarchical diagnostic; \textit{LGame} = LangGame
      \item (language-agnostic control); \textit{MDA} = Multi-Digit Addition (catastrophic-forgetting probe); CUTE = character-level 
      \item understanding control). Bold = best in column. Cell color: \colorbox[HTML]{F4A582}{\strut low} $\to$ white $\to$ \colorbox[HTML]{92C5DE}{high} per column.
    \end{tablenotes}
  \end{threeparttable}
\end{table*}

% ---- Commercial model results (GEN only) ----
\begin{table*}[htbp]
  \centering
  \caption{Results for commercial models on PACUTE GEN
    benchmarks. EM = exact match; CM = contains match.
    Commercial APIs do not expose token log-probabilities, so
    MCQ benchmarks are omitted. All values are percentages.}
  \label{tab:results-commercial}
  \begin{threeparttable}
    \begin{adjustbox}{max width=\textwidth}
      \begin{tabular}{lrrrrrrrrrrrrrrrrrr}
\toprule
\textbf{Model} & \multicolumn{2}{c}{\textbf{Comp-Gen}} & \multicolumn{2}{c}{\textbf{Manip-Gen}} & \multicolumn{2}{c}{\textbf{MExt-Gen}} & \multicolumn{2}{c}{\textbf{MProd-Gen}} & \multicolumn{2}{c}{\textbf{Syll-Gen}} & \multicolumn{2}{c}{\textbf{Hier-Gen}} & \multicolumn{2}{c}{\textbf{LGame-Gen}} & \multicolumn{2}{c}{\textbf{MDA-Gen}} & \multicolumn{2}{c}{\textbf{CUTE-Gen}} \\
\cmidrule(lr){2-3}\cmidrule(lr){4-5}\cmidrule(lr){6-7}\cmidrule(lr){8-9}\cmidrule(lr){10-11}\cmidrule(lr){12-13}\cmidrule(lr){14-15}\cmidrule(lr){16-17}\cmidrule(lr){18-19}
 & \textit{EM} & \textit{CM} & \textit{EM} & \textit{CM} & \textit{EM} & \textit{CM} & \textit{EM} & \textit{CM} & \textit{EM} & \textit{CM} & \textit{EM} & \textit{CM} & \textit{EM} & \textit{CM} & \textit{EM} & \textit{CM} & \textit{EM} & \textit{CM} \\
\midrule
claude-haiku-4-5                       & \cellcolor[HTML]{94C6DE}99.8           & \cellcolor[HTML]{92C5DE}\textbf{100.0} & \cellcolor[HTML]{FDF4F1}82.0           & \cellcolor[HTML]{FDF2ED}82.0           & \cellcolor[HTML]{FDF1EC}59.75 &
  \cellcolor[HTML]{FDF5F1}64.0 &
  \cellcolor[HTML]{F9CFBC}45.3 &
  \cellcolor[HTML]{F9CEBB}45.3 & \cellcolor[HTML]{CAE3EF}69.8          & \cellcolor[HTML]{B9DAEA}74.2          & \cellcolor[HTML]{FDF4F0}52.5          & \cellcolor[HTML]{C3DFEC}71.8          & \cellcolor[HTML]{9ECBE1}99.7           & \cellcolor[HTML]{9ECBE1}99.7           & \cellcolor[HTML]{92C5DE}\textbf{100.0} & \cellcolor[HTML]{92C5DE}\textbf{100.0} & \cellcolor[HTML]{BEDCEB}88.4          & \cellcolor[HTML]{EDF5F9}89.1          \\
        claude-sonnet-4-6                      & \cellcolor[HTML]{FEF7F4}90.4           & \cellcolor[HTML]{FEF7F4}90.4           & \cellcolor[HTML]{B9DAE9}94.1           & \cellcolor[HTML]{B9DAEA}94.2           & \cellcolor[HTML]{DDEDF4}74.0          & \cellcolor[HTML]{E5F1F7}76.2          & \cellcolor[HTML]{FBE1D5}54.0           & \cellcolor[HTML]{FBE1D5}54.0          & \cellcolor[HTML]{FDFEFE}58.0          & \cellcolor[HTML]{F8FBFC}60.5          & \cellcolor[HTML]{DCECF4}57.2          & \cellcolor[HTML]{E4F1F7}68.0          & \cellcolor[HTML]{96C7DF}99.9           & \cellcolor[HTML]{96C7DF}99.9           & \cellcolor[HTML]{92C5DE}\textbf{100.0}          & \cellcolor[HTML]{92C5DE}\textbf{100.0}          & \cellcolor[HTML]{AFD4E7}91.9          & \cellcolor[HTML]{C2DFEC}93.9          \\
        claude-opus-4-6                        & \cellcolor[HTML]{92C5DE}\textbf{100.0} & \cellcolor[HTML]{92C5DE}\textbf{100.0}          & \cellcolor[HTML]{9CCAE1}98.5           & \cellcolor[HTML]{9AC9E0}98.8           & \cellcolor[HTML]{C9E2EE}77.3 &
  \cellcolor[HTML]{BFDDEB}80.3 &
  \cellcolor[HTML]{C6E1ED}63.3 &
  \cellcolor[HTML]{C2DFEC}64.0 & \cellcolor[HTML]{B1D5E7}75.6          & \cellcolor[HTML]{A5CFE4}78.6          & \cellcolor[HTML]{EBF4F9}55.7          & \cellcolor[HTML]{9DCBE1}76.2          & \cellcolor[HTML]{96C7DF}99.9           & \cellcolor[HTML]{96C7DF}99.9           & \cellcolor[HTML]{92C5DE}\textbf{100.0}          & \cellcolor[HTML]{92C5DE}\textbf{100.0}          & \cellcolor[HTML]{9CCAE1}96.6          & \cellcolor[HTML]{9ECBE1}97.9          \\
deepseek/deepseek-r1-0528 (thinking) & \cellcolor[HTML]{FCEEE8}89.5 & \cellcolor[HTML]{FCEEE8}89.5 & \cellcolor[HTML]{A7D0E4}96.9 & \cellcolor[HTML]{A7D0E4}96.9 & \cellcolor[HTML]{D6E9F2}75.5 & \cellcolor[HTML]{E6F2F7}76.0 & \cellcolor[HTML]{FEFBFA}62.0 & \cellcolor[HTML]{FEFBFA}62.0 & \cellcolor[HTML]{FDF5F1}55.0 & \cellcolor[HTML]{FDF0EA}55.0 & \cellcolor[HTML]{BDDCEB}60.3 & \cellcolor[HTML]{BCDBEA}72.7 & \cellcolor[HTML]{92C5DE}\textbf{100.0} & \cellcolor[HTML]{92C5DE}\textbf{100.0} & \cellcolor[HTML]{92C5DE}\textbf{100.0} & \cellcolor[HTML]{92C5DE}\textbf{100.0} & \cellcolor[HTML]{E2F0F6}79.3 & \cellcolor[HTML]{F9CFBC}80.8 \\
deepseek/deepseek-v3.2-exp (thinking) & \cellcolor[HTML]{FCE8E0}88.9 & \cellcolor[HTML]{FCE8E0}88.9 & \cellcolor[HTML]{FCFDFE}84.1 & \cellcolor[HTML]{FEFEFE}84.4 & \cellcolor[HTML]{DBEBF4}74.5 & \cellcolor[HTML]{E1EFF5}77.0 & \cellcolor[HTML]{FCE7DE}56.0 & \cellcolor[HTML]{FCE7DE}56.0 & \cellcolor[HTML]{F7C3AC}41.0 & \cellcolor[HTML]{F7C1A9}42.5 & \cellcolor[HTML]{FEFCFC}53.5 & \cellcolor[HTML]{EAF4F8}67.3 & \cellcolor[HTML]{9ECBE1}99.7 & \cellcolor[HTML]{9ECBE1}99.7 & \cellcolor[HTML]{B6D8E9}99.9 & \cellcolor[HTML]{B6D8E9}99.9 & \cellcolor[HTML]{B7D8E9}90.1 & \cellcolor[HTML]{E1EFF6}90.4 \\
google/gemini-3.1-flash-lite & \cellcolor[HTML]{F9D2C1}86.7 & \cellcolor[HTML]{F9D2C1}86.7 & \cellcolor[HTML]{FEFBF9}83.1 & \cellcolor[HTML]{FEFCFB}83.8 & \cellcolor[HTML]{C2DEEC}79.5 & \cellcolor[HTML]{CCE4EF}80.5 & \cellcolor[HTML]{FDFEFE}63.3 & \cellcolor[HTML]{FDFEFE}63.3 & \cellcolor[HTML]{F5B395}36.5 & \cellcolor[HTML]{F5B091}38.0 & \cellcolor[HTML]{FDF4F0}52.5 & \cellcolor[HTML]{F4F9FB}66.2 & \cellcolor[HTML]{ABD2E5}99.4 & \cellcolor[HTML]{ABD2E5}99.4 & \cellcolor[HTML]{92C5DE}\textbf{100.0} & \cellcolor[HTML]{92C5DE}\textbf{100.0} & \cellcolor[HTML]{E2EFF6}79.4 & \cellcolor[HTML]{E3F0F6}90.2 \\
google/gemini-3.5-flash (thinking) & \cellcolor[HTML]{FEF7F4}90.4 & \cellcolor[HTML]{FEF7F4}90.4 & \cellcolor[HTML]{9DCBE1}98.2 & \cellcolor[HTML]{9ECBE1}98.2 & \cellcolor[HTML]{92C5DE}\textbf{89.5} & \cellcolor[HTML]{92C5DE}\textbf{90.5} & \cellcolor[HTML]{92C5DE}\textbf{90.0} & \cellcolor[HTML]{92C5DE}\textbf{90.0} & \cellcolor[HTML]{94C6DE}82.5 & \cellcolor[HTML]{94C6DE}82.5 & \cellcolor[HTML]{F8C8B3}47.2 & \cellcolor[HTML]{FCEAE2}62.2 & \cellcolor[HTML]{AFD4E7}99.3 & \cellcolor[HTML]{AFD4E7}99.3 & \cellcolor[HTML]{B6D8E9}99.9 & \cellcolor[HTML]{B6D8E9}99.9 & \cellcolor[HTML]{F4A582}45.4 & \cellcolor[HTML]{9ECBE1}97.9 \\
gpt-5-mini & \cellcolor[HTML]{FEF7F4}90.4 & \cellcolor[HTML]{FEF7F4}90.4 & \cellcolor[HTML]{95C6DF}99.5 & \cellcolor[HTML]{95C6DF}99.5 & \cellcolor[HTML]{DBEBF4}74.5 & \cellcolor[HTML]{EBF4F9}75.2 & \cellcolor[HTML]{FEF7F4}60.7 & \cellcolor[HTML]{FEF7F4}60.7 & \cellcolor[HTML]{F9CFBD}44.5 & \cellcolor[HTML]{F9CEBB}46.0 & \cellcolor[HTML]{F0F7FA}55.2 & \cellcolor[HTML]{E7F2F7}67.7 & \cellcolor[HTML]{92C5DE}\textbf{100.0} & \cellcolor[HTML]{92C5DE}\textbf{100.0} & \cellcolor[HTML]{92C5DE}\textbf{100.0} & \cellcolor[HTML]{92C5DE}\textbf{100.0} & \cellcolor[HTML]{B9D9E9}89.6 & \cellcolor[HTML]{97C7DF}98.7 \\
gpt-5.4-nano & \cellcolor[HTML]{F4A582}82.2 & \cellcolor[HTML]{F4A582}82.2 & \cellcolor[HTML]{F4A582}67.6 & \cellcolor[HTML]{F4A582}68.5 & \cellcolor[HTML]{F4A582}44.8 & \cellcolor[HTML]{F4A582}53.2 & \cellcolor[HTML]{F4A582}36.0 & \cellcolor[HTML]{F4A582}36.0 & \cellcolor[HTML]{F4A582}32.5 & \cellcolor[HTML]{F4A582}35.0 & \cellcolor[HTML]{F4A582}42.8 & \cellcolor[HTML]{F4A582}52.5 & \cellcolor[HTML]{F4A582}94.9 & \cellcolor[HTML]{F4A582}94.9 & \cellcolor[HTML]{F4A582}99.4 & \cellcolor[HTML]{F4A582}99.4 & \cellcolor[HTML]{F6FAFC}74.4 & \cellcolor[HTML]{F4A582}75.1 \\
gpt-5.4-mini & \cellcolor[HTML]{F8CBB7}86.0 & \cellcolor[HTML]{F8CDBA}86.2 & \cellcolor[HTML]{FADCCF}77.6 & \cellcolor[HTML]{FADBCD}78.0 & \cellcolor[HTML]{FBE5DB}60.8 & \cellcolor[HTML]{FADDD0}65.0 & \cellcolor[HTML]{FADCCF}52.7 & \cellcolor[HTML]{FADCCF}52.7 & \cellcolor[HTML]{F7C3AC}41.0 & \cellcolor[HTML]{F6BBA1}41.0 & \cellcolor[HTML]{E8F3F8}56.0 & \cellcolor[HTML]{D6E9F2}69.7 & \cellcolor[HTML]{D2E7F1}98.5 & \cellcolor[HTML]{D2E7F1}98.5 & \cellcolor[HTML]{FFFFFF}99.7 & \cellcolor[HTML]{FFFFFF}99.7 & \cellcolor[HTML]{C2DEEC}87.4 & \cellcolor[HTML]{FAFCFD}87.6 \\
gpt-5.5 & \cellcolor[HTML]{FEFDFC}90.9 & \cellcolor[HTML]{FEFDFC}90.9 & \cellcolor[HTML]{92C5DE}\textbf{100.0} & \cellcolor[HTML]{92C5DE}\textbf{100.0} & \cellcolor[HTML]{CEE5F0}77.0 & \cellcolor[HTML]{DCECF4}77.8 & \cellcolor[HTML]{B2D6E7}82.0 & \cellcolor[HTML]{B2D6E7}82.0 & \cellcolor[HTML]{92C5DE}\textbf{83.0} & \cellcolor[HTML]{92C5DE}\textbf{83.0} & \cellcolor[HTML]{92C5DE}\textbf{64.7} & \cellcolor[HTML]{92C5DE}\textbf{77.5} & \cellcolor[HTML]{92C5DE}\textbf{100.0} & \cellcolor[HTML]{92C5DE}\textbf{100.0} & \cellcolor[HTML]{92C5DE}\textbf{100.0} & \cellcolor[HTML]{92C5DE}\textbf{100.0} & \cellcolor[HTML]{92C5DE}\textbf{99.3} & \cellcolor[HTML]{92C5DE}\textbf{99.3} \\
moonshotai/kimi-k2-thinking (thinking) & \cellcolor[HTML]{FCE8E0}88.9 & \cellcolor[HTML]{FCE8E0}88.9 & \cellcolor[HTML]{B5D7E8}94.8 & \cellcolor[HTML]{B6D8E9}94.8 & \cellcolor[HTML]{E1EFF5}73.2 & \cellcolor[HTML]{EFF6FA}74.5 & \cellcolor[HTML]{FDF5F1}60.0 & \cellcolor[HTML]{FDF5F1}60.0 & \cellcolor[HTML]{FDF1EC}54.0 & \cellcolor[HTML]{FEFDFC}58.5 & \cellcolor[HTML]{E3F0F6}56.5 & \cellcolor[HTML]{ECF4F9}67.2 & \cellcolor[HTML]{ABD2E5}99.4 & \cellcolor[HTML]{A3CEE3}99.6 & \cellcolor[HTML]{B6D8E9}99.9 & \cellcolor[HTML]{B6D8E9}99.9 & \cellcolor[HTML]{BBDAEA}89.1 & \cellcolor[HTML]{BFDCEB}94.3 \\
        \bottomrule
      \end{tabular}
    \end{adjustbox}
    \begin{tablenotes}[flushleft]\footnotesize
      \item \textit{Comp} = PACUTE Composition; \textit{Manip} = PACUTE Manipulation; \textit{MExt} = PACUTE Morphological Extraction; \textit{MProd} = 
      \item PACUTE Morphological Production; \textit{Syll} = PACUTE Syllabification; \textit{Hier} = Hierarchical diagnostic; \textit{LGame} = LangGame
      \item (language-agnostic control); \textit{MDA} = Multi-Digit Addition (catastrophic-forgetting probe); CUTE = character-level 
      \item understanding control). Bold = best in column. Cell color: \colorbox[HTML]{F4A582}{\strut low} $\to$ white $\to$ \colorbox[HTML]{92C5DE}{high} per column.
    \end{tablenotes}
  \end{threeparttable}
\end{table*}

\clearpage
\section{Continued Pre-training Results}
\label{sec:cpt-results}

Results from continued pre-training of Gemma-2-2B on SEA-PILE v2 Filipino corpus ($\sim$7.4GB) across three tokenization regimes: Vanilla (standard BPE), StochasTok ($\sim$10\% stochastic token expansion), and Patok (morphology-aware expand/contract). MCQ chance = 25\% (4-way). All CPT checkpoints lose math capability entirely (catastrophic forgetting), confirming the multi-digit addition benchmark as a reliable domain-adaptation health check.

\begin{table*}[htbp]
    \centering
    \small
    \caption{MCQ benchmark accuracy for Gemma-2-2B and its
  three CPT regimes on SEA-PILE v2 (5,000 steps). All values
   are raw 4-way accuracy percentages (chance = 25\%). Bold
  = best per column. CPT regimes broadly underperform the
  un-CPT'd base on character-level tasks and the
  LangGame/MDA controls, with MDA in particular dropping
  from 68.7\% to 14--18\% (catastrophic forgetting on
  arithmetic).}
    \label{tab:cpt-results}
    \begin{threeparttable}
      \begin{adjustbox}{max width=\textwidth}
        \begin{tabular}{lrrrrrrrrr}
          \toprule
          \textbf{Regime} & \textbf{Comp-MCQ} & \textbf{Manip-MCQ} & \textbf{MExt-MCQ} & \textbf{MProd-MCQ} & \textbf{Syll-MCQ} &\textbf{PACUTE MCQ Avg.} & \textbf{Hier-MCQ} & \textbf{LGame-MCQ} & \textbf{MDA-MCQ} \\
          \midrule
          Base (no CPT)   & \textbf{36.6} & 27.3 &
  \textbf{66.5} & 82.7 & \textbf{32.0} & \textbf{49.0} &
  23.5 & \textbf{41.0} & \textbf{68.7} \\
          \midrule
          Vanilla BPE     & 25.8 & 34.9 & 57.5 &
  \textbf{85.3} & 21.0 & 44.9 & 22.5 & 28.0 & 17.4 \\
          StochasTok      & 20.1 & 32.5 & 50.7 & 78.0 & 23.5
   & 41.0 & 21.2 & 28.2 & 14.8 \\
          Patok           & 27.6 & \textbf{35.2} & 54.3 &
  78.0 & 19.5 & 42.9 & \textbf{24.7} & 27.1 & 17.9 \\
          \bottomrule
        \end{tabular}
      \end{adjustbox}
      \begin{tablenotes}[flushleft]\footnotesize
        \item \textit{Comp} = PACUTE Composition; \textit{Manip} = PACUTE Manipulation; \textit{MExt} = PACUTE Morphological Extraction; \textit{MProd} = 
        \item PACUTE Morphological Production; \textit{Syll} = PACUTE Syllabification; \textit{PACUTE MCQ Avg.} is the macro-average of the 
        \item PACUTE MCQ main suite; \textit{Hier} = Hierarchical diagnostic; \textit{LGame} = LangGame (language-agnostic control); \textit{MDA} = Multi-
        \item Digit Addition (catastrophic-forgetting probe); Bold = best in column.
      \end{tablenotes}
    \end{threeparttable}
  \end{table*}

\section{Error Analysis: Annotated Examples}
\label{sec:error-examples}

Tables~\ref{tab:error-type1}--\ref{tab:error-type4} provide representative model outputs for each of the four error types described in \S\ref{sec:error-analysis}. All examples are drawn from GEN inference outputs of instruction-tuned models. Task inputs are quoted verbatim; model outputs are lightly formatted for readability (angle-bracket tags are from the model's structured output format).

\begin{table*}[ht]
  \centering
  \caption{Type I: Instruction following failure. Model produces a malformed output---here, a character-insertion task yields garbled output with spurious spaces rather than the target string.}
  \label{tab:error-type1}
  \small
  \begin{tabular}{p{0.45\linewidth} p{0.45\linewidth}}
    \toprule
    \textbf{Input}                                                                      & \textbf{Model output (Qwen3-8B)}                                    \\
    \midrule
    \textit{Put a character ``b'' after every character ``o'' in the word ``tungkol''.} &
    \texttt{<reflection>}I identified the operation as inserting ``b'' after each ``o''.\texttt{</reflection>} \texttt{<answer>}tu ngbk b l\texttt{</answer>} \\
    \midrule
    Gold answer: \textit{tungkobl}                                                      & (exact match: \ding{55}; contains match: \ding{55})                 \\
    \bottomrule
  \end{tabular}
\end{table*}

\begin{table*}[ht]
  \centering
  \caption{Type II: Linguistic error. The model correctly identifies that \textit{-in-} is an infix and that it should be placed inside the root, but inserts it at the wrong position (\textit{tak\textbf{in}bo} vs.\ the correct \textit{\textbf{tin}akbo}, where the infix follows the initial consonant).}
  \label{tab:error-type2}
  \small
  \begin{tabular}{p{0.45\linewidth} p{0.45\linewidth}}
    \toprule
    \textbf{Input}                                                 & \textbf{Model output (Gemma-3-27B-IT)}                                                                            \\
    \midrule
    \textit{Inflect the word ``takbo'' to use the infix ``-in-''.} &
    \texttt{<reflection>}Infixation inserts an affix within the root word; \textit{-in-} results in \textit{takinbo}.\texttt{</reflection>} \texttt{<answer>}takinbo\texttt{</answer>} \\
    \midrule
    Gold answer: \textit{tinakbo}                                  & (exact match: \ding{55}; contains match: \ding{55})                                                               \\
    \bottomrule
  \end{tabular}
\end{table*}

\begin{table*}[ht]
  \centering
  \caption{Type III: Reasoning inconsistency. The model's chain-of-thought correctly derives \textit{kinain} (``k + in + ain'') but then dismisses this intermediate result and outputs \textit{kalinin} as the final answer.}
  \label{tab:error-type3}
  \small
  \begin{tabular}{p{0.45\linewidth} p{0.45\linewidth}}
    \toprule
    \textbf{Input}                                                & \textbf{Model output (Qwen3-8B-Thinking)}           \\
    \midrule
    \textit{Inflect the word ``kain'' to use the infix ``-in-''.} &
    \texttt{<reflection>}\ldots\ inserting \textit{-in-} between `k' and `a': ``k'' + ``in'' + ``ain''? that would be \textit{kinain}? wait, but \textit{kinain} is the past tense \ldots\ maybe the correct answer is \textit{kalinin}.\texttt{</reflection>}
    \texttt{<answer>}kalinin\texttt{</answer>}                                                                          \\
    \midrule
    Gold answer: \textit{kinain}                                  & (exact match: \ding{55}; contains match: \ding{55}) \\
    \bottomrule
  \end{tabular}
\end{table*}

\begin{table*}[ht]
  \centering
  \caption{Type IV: Systematic penultimate-stress bias. The model explicitly asserts that Filipino stress defaults to the penultimate syllable and selects \textit{ka} accordingly. In the given sentence, \textit{kaya} means ``which is why'' (\textit{kayà}), which carries final stress on \textit{ya}.}
  \label{tab:error-type4}
  \small
  \begin{tabular}{p{0.45\linewidth} p{0.45\linewidth}}
    \toprule
    \textbf{Input}                                                                                                                         & \textbf{Model output (Gemma-3-27B-IT)}                                                                                                                    \\
    \midrule
    \textit{Based on the sentence ``Ito ang unang beses ko, kaya ako kabada.'', which syllable of the word ``kaya'' has the stress?} &
    \texttt{<reflection>}the sentence indicates ``kaya'' means ``can'' or ``able to'', implying a verb usage. in filipino, verbs generally have stress on the penultimate (second to last) syllable.\texttt{</reflection>}\texttt{<answer>}ka\texttt{</answer>} \\
    \midrule
    Gold answer: \textit{ya}                                                                                                               & (exact match: \ding{55}; contains match: \ding{55})                                                                                                       \\
    \bottomrule
  \end{tabular}
\end{table*}

\end{document}

%% file: fig_results.tex
 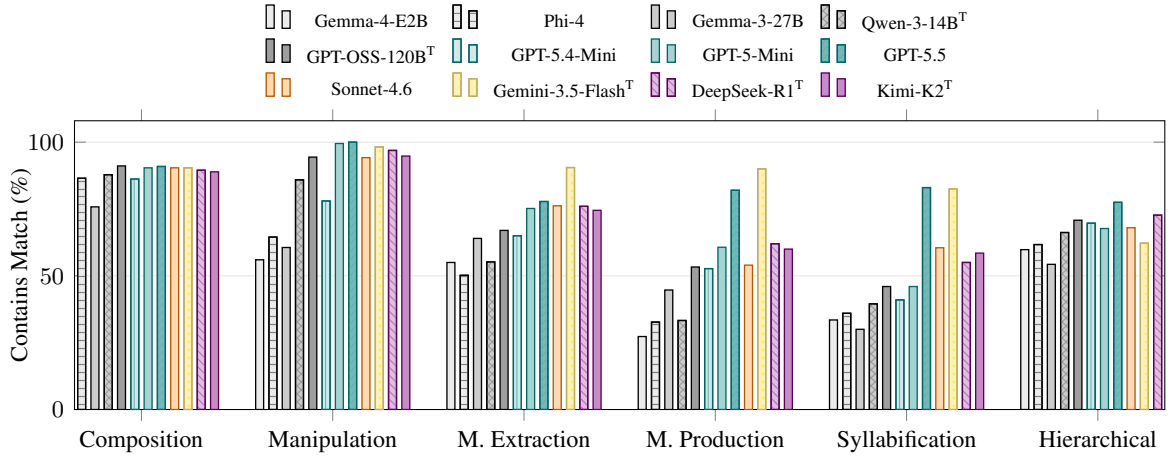
\begin{figure*}[t]
  \centering
  \begin{tikzpicture}
  \begin{axis}[
      ybar,
      width=\textwidth,
      height=5.4cm,
      ymin=0, ymax=108,
      ylabel={Contains Match (\%)},
      ylabel style={font=\small, at={(axis description
  cs:-0.03,0.5)}},
      yticklabel style={font=\footnotesize},
      xticklabel style={font=\small},
      ymajorgrids=true,
      grid style={gray!20},
      bar width=3.0pt,
      enlarge x limits=0.07,
      xtick=data,
      symbolic x coords={Composition, Manipulation, M. 
  Extraction, M. Production, Syllabification,
  Hierarchical},
      legend style={
          font=\scriptsize, at={(0.5,1.03)}, anchor=south,
          legend columns=4, column sep=4pt, draw=none,
  fill=none,
      },
      every outer y axis line/.append style={black},
      every outer x axis line/.append style={black},
  ]

  % ===== Open-weight (gray) — selected from newer eval pool
   with MExt/MProd data =====
  % gemma-4-E2B-it (~5B)
  \addplot[fill=black!8, draw=black, line width=0.5pt]
      coordinates {(Composition, 70.9) (Manipulation, 56.0)
  (M. Extraction, 55.0) (M. Production, 27.3) (Syllabification, 33.5) (Hierarchical, 59.8)};
  % Phi-4 (14B)
  \addplot[fill=black!8, draw=black, line width=0.5pt,
           postaction={pattern=horizontal lines, pattern
  color=black!50}]
      coordinates {(Composition, 86.5) (Manipulation, 64.5) (M. Extraction, 50.2) (M. Production, 32.7)(Syllabification, 36.0) (Hierarchical, 61.7)};
  % gemma-3-27B-it (27B)
  \addplot[fill=black!20, draw=black, line width=0.5pt]
      coordinates {(Composition, 75.8) (Manipulation, 60.6)
  (M. Extraction, 64.0) (M. Production, 44.7)
  (Syllabification, 30.0) (Hierarchical, 54.3)};
  % Qwen3-14B (thinking)
  \addplot[fill=black!20, draw=black, line width=0.5pt,
           postaction={pattern=crosshatch, pattern
  color=black!40}]
      coordinates {(Composition, 87.8) (Manipulation, 85.9)
  (M. Extraction, 55.2) (M. Production, 33.3)
  (Syllabification, 39.5) (Hierarchical, 66.2)};
  % gpt-oss-120b (thinking)
  \addplot[fill=black!38, draw=black, line width=0.5pt]
      coordinates {(Composition, 91.1) (Manipulation, 94.4)
  (M. Extraction, 67.0) (M. Production, 53.3)
  (Syllabification, 46.0) (Hierarchical, 70.8)};

  % ===== Commercial (color) =====
  % gpt-5.4-mini
  \addplot[fill=teal!18, draw=teal!80!black, line
  width=0.5pt,
           postaction={pattern=vertical lines, pattern
  color=teal!60}]
      coordinates {(Composition, 86.2) (Manipulation, 78.0)
  (M. Extraction, 65.0) (M. Production, 52.7)
  (Syllabification, 41.0) (Hierarchical, 69.7)};
  % gpt-5-mini
  \addplot[fill=teal!35, draw=teal!80!black, line
  width=0.5pt]
      coordinates {(Composition, 90.4) (Manipulation, 99.5)
  (M. Extraction, 75.2) (M. Production, 60.7)
  (Syllabification, 46.0) (Hierarchical, 67.7)};
  % gpt-5.5
  \addplot[fill=teal!55, draw=teal!80!black, line
  width=0.5pt,
           postaction={pattern=north east lines, pattern
  color=teal!60}]
      coordinates {(Composition, 90.9) (Manipulation, 100.0)
   (M. Extraction, 77.8) (M. Production, 82.0)
  (Syllabification, 83.0) (Hierarchical, 77.5)};
  % Claude Sonnet 4.6
  \addplot[fill=orange!30, draw=orange!70!black, line
  width=0.5pt]
      coordinates {(Composition, 90.4) (Manipulation, 94.2)
  (M. Extraction, 76.2) (M. Production, 54.0)
  (Syllabification, 60.5) (Hierarchical, 68.0)};
  % Gemini-3.5-Flash (thinking)
  \addplot[fill=Goldenrod!40, draw=Goldenrod!70!black, line
  width=0.5pt,
           postaction={pattern=dots, pattern
  color=Goldenrod!70}]
      coordinates {(Composition, 90.4) (Manipulation, 98.2)
  (M. Extraction, 90.5) (M. Production, 90.0)
  (Syllabification, 82.5) (Hierarchical, 62.2)};
  % DeepSeek-R1 (reasoning)
  \addplot[fill=violet!25, draw=violet!70!black, line
  width=0.5pt,
           postaction={pattern=north west lines, pattern
  color=violet!55}]
      coordinates {(Composition, 89.5) (Manipulation, 96.9)
  (M. Extraction, 76.0) (M. Production, 62.0)
  (Syllabification, 55.0) (Hierarchical, 72.7)};
  % Kimi-K2-Thinking
  \addplot[fill=violet!45, draw=violet!70!black, line
  width=0.5pt]
      coordinates {(Composition, 88.9) (Manipulation, 94.8)
  (M. Extraction, 74.5) (M. Production, 60.0)
  (Syllabification, 58.5) (Hierarchical, 67.2)};

  \legend{
      Gemma-4-E2B, Phi-4, Gemma-3-27B,
  Qwen-3-14B\textsuperscript{T},
  GPT-OSS-120B\textsuperscript{T},
      GPT-5.4-Mini, GPT-5-Mini, GPT-5.5, Sonnet-4.6,
  Gemini-3.5-Flash\textsuperscript{T},
      DeepSeek-R1\textsuperscript{T},
  Kimi-K2\textsuperscript{T}
  }
  \end{axis}
  \end{tikzpicture}
  \caption{\textbf{Contains-match accuracy across PACUTE
  categories} for 12 models spanning open-weight (gray,
  5B--120B) and frontier commercial (color) tiers.
  Superscript \textsuperscript{T}\,=\,extended
  thinking/reasoning. Composition and Manipulation approach
  ceiling for strong models, while \textbf{Hierarchical and
  Syllabification remain suppressed across all classes}.
  Even GPT-5.5 (77.5\% Hierarchical-CM) sits 22pp below its
  Manipulation ceiling. DeepSeek-R1 leads Hierarchical
  (72.7\% CM); Gemini-3.5-Flash dominates Morphological
  Extraction and Production (90.5/90.0\% CM) but flops on
  Hierarchical (62.2\%), suggesting morpheme operations and
  multi-step composition recruit partially disjoint
  capabilities.}
  \label{fig:results_by_category}
  \end{figure*}